\documentclass[lettersize,journal]{IEEEtran}
\usepackage{amsmath,amsfonts}
\usepackage[T1]{fontenc}
\usepackage{rotating}
\usepackage{array}
\usepackage{textcomp}
\usepackage[ruled,vlined,onelanguage]{algorithm2e} 
\usepackage{algpseudocode}
\usepackage[table,xcdraw]{xcolor}
\usepackage{stfloats}
\usepackage{url}
\usepackage{verbatim}
\usepackage{graphicx}
\usepackage{longtable}
\usepackage{academicons}
\usepackage{multirow}
\newcommand{\orcid}[1]{\href{https://orcid.org/#1}{\textcolor[HTML]{A6CE39}{\aiOrcid}}}
\usepackage{tikz}
\usetikzlibrary{svg.path}
\usepackage[colorlinks=true, pdfstartview=FitV, linkcolor=blue, citecolor=blue, urlcolor=blue]{hyperref}
\usepackage{lscape}
\usepackage{cite}
\usepackage{amsmath}
\usepackage{amssymb}
\usepackage{xcolor}
\usepackage{caption}
\usepackage{xr}
\usepackage{orcidlink}
\usepackage{booktabs,caption}
\usepackage[flushleft]{threeparttable}

\usepackage[colorlinks=true, pdfstartview=FitV, linkcolor=blue, citecolor=blue, urlcolor=blue]{hyperref}

\begin{document}

\title{Improving Trajectory Prediction in Dynamic Multi-Agent Environment by Dropping Waypoints}

\author{Pranav Singh Chib$^{\orcidlink{0000-0003-4930-3937}}$, \and Pravendra Singh$^{\orcidlink{0000-0003-1001-2219}}$


\thanks{(Corresponding author: Pravendra Singh.)}
\thanks{Pranav Singh Chib and Pravendra Singh are with the Department of Computer Science and Engineering, Indian Institute of Technology Roorkee, Uttarakhand 247667, India,
(e-mail: pranavs\_chib@cs.iitr.ac.in; pravendra.singh@cs.iitr.ac.in).}}



\maketitle

\begin{abstract}
The inherently diverse and uncertain nature of trajectories presents a formidable challenge in accurately modeling them. Motion prediction systems must effectively learn spatial and temporal information from the past to forecast the future trajectories of the agent. Many existing methods learn temporal motion via separate components within stacked models to capture temporal features. Furthermore, prediction methods often operate under the assumption that observed trajectory waypoint sequences are complete, disregarding scenarios where missing values may occur, which can influence their performance. Moreover, these models may be biased toward particular waypoint sequences when making predictions. We propose a novel approach called \textit{Temporal Waypoint Dropping (TWD)} that explicitly incorporates temporal dependencies during the training of a trajectory prediction model. By stochastically dropping waypoints from past observed trajectories, the model is forced to learn the underlying temporal representation from the remaining waypoints, resulting in an improved model. Incorporating stochastic temporal waypoint dropping into the model learning process significantly enhances its performance in scenarios with missing values. Experimental results demonstrate our approach's substantial improvement in trajectory prediction capabilities. Our approach can complement existing trajectory prediction methods to improve their prediction accuracy. We evaluate our proposed approach on three datasets: NBA Sports VU, ETH-UCY, and TrajNet++.
\end{abstract}

\begin{IEEEkeywords}
Trajectory prediction, waypoints dropping, intelligent vehicles, neural network, deep learning.
\end{IEEEkeywords}

\section{Introduction}
\IEEEPARstart{T}{rajectory} prediction refers to forecasting the future paths of one or more agents, given their historical movement patterns, and it holds significant importance for various fields, such as driverless cars, drone technology, security monitoring systems, robotics, and human-robot collaboration. Multi-agent trajectory prediction deals with two primary aspects: the time dimension, where we analyze how previous agent states impact their future states, and the social dimension, which considers how the actions of one agent influence those of other agents. The prediction task is often a continuous streaming task where the current temporal state evolves into a historical state over time. The historical state typically influences the successive predicted trajectories. Thus, ensuring the temporal dependency becomes critical for downstream predictions. Significant efforts \cite{shi2021sgcn, mohamed2020social, sekhon2021scan} have been made in temporal and social modeling over the past few years. 
\begin{figure*}
    \centering
    \includegraphics[scale=.142]{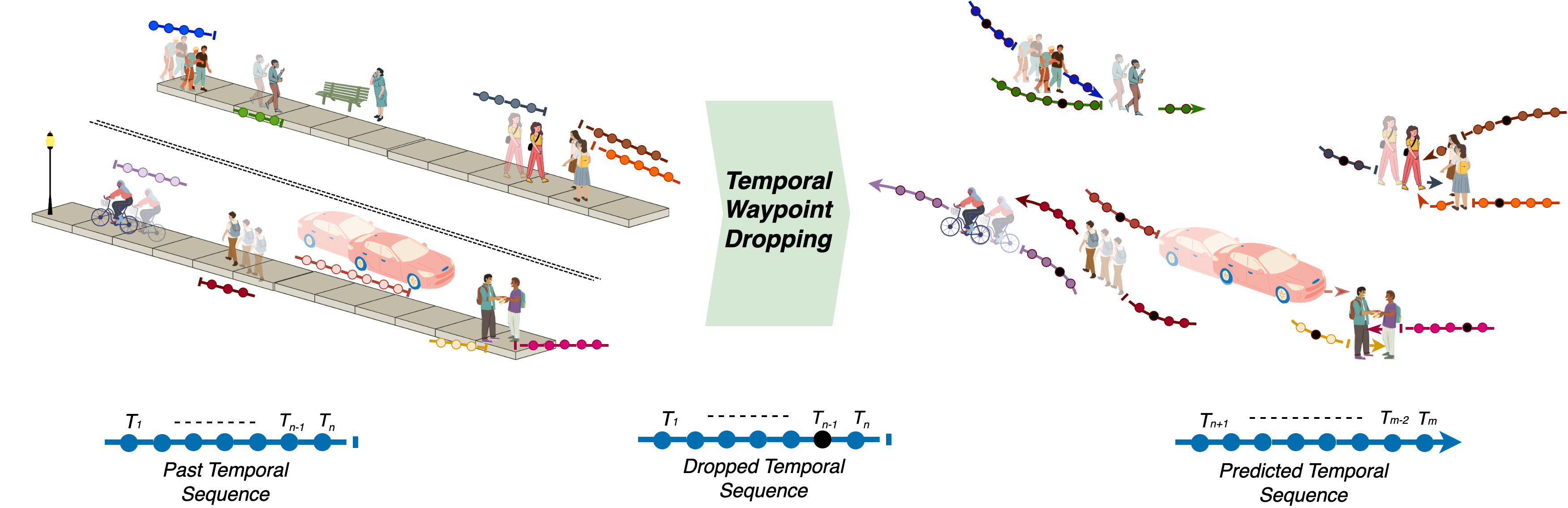}
    \caption{Visualizing trajectory prediction in a dynamically changing social environment, where the temporal drop approach positively impacts precise future trajectory predictions. The faded view represents the previous temporal motion, and the black waypoint indicates the dropped waypoint.}
    \label{final}
\end{figure*}

Many existing trajectory prediction approaches \cite{shi2021sgcn, mohamed2020social, sekhon2021scan,gupta2018social} employ separate models to learn temporal motion and then feed the trajectories into stacked models of temporal components to extract temporal features. Specifically, these temporal models utilize sequential networks, such as Recurrent Neural Networks \cite{cao2018brits} and their variations, Temporal Convolutional Networks \cite{bai2018empirical}, Self-attention mechanisms \cite{vaswani2017attention}, and Transformers \cite{vaswani2017attention} to capture temporal dependencies effectively. In this paper, we propose \textit{Temporal Waypoint Dropping (TWD)}, a novel approach that involves dropping temporal waypoints in past observed trajectories. We explicitly incorporate temporal dependencies during the training of a trajectory prediction model by dropping waypoints from past observed trajectories to improve the model's understanding of temporal dependencies. Empirical results demonstrate that our approach significantly improves the model's capabilities in trajectory prediction.

Trajectory prediction models may be biased towards specific waypoint sequences, leading to a strong reliance on these particular sequences for future predictions, which may limit their trajectory prediction performance. To address this issue, our approach, Temporal Waypoint Dropping, utilizes a stochastic waypoint dropping to reduce such bias. By stochastically dropping waypoints from past observed trajectories, the model is compelled to learn the underlying temporal representation from the remaining waypoints, resulting in an improved model's understanding of the underlying temporal representation. We empirically demonstrate that stochastically dropping waypoints during training improves trajectory prediction performance.

Furthermore, existing trajectory prediction methods typically assume that observed trajectory waypoint sequences are complete, ignoring scenarios where missing values may occur. This assumption can hinder prediction performance, particularly in situations with incomplete data, as models are designed with the assumption of completeness, negatively affecting model predictions, as demonstrated in ablation experiments (Section~\ref{missing}). To overcome this limitation, we introduce stochastic temporal waypoint dropping into the model learning process. This allows the model to adapt these scenarios better, leading to significant improvements in prediction accuracy as compared to the baseline model (see Section~\ref{missing}).

Our approach (Figure~\ref{final}) can be easily integrated with existing trajectory prediction methods. We integrated our approach into three existing methods: the generative-based GroupNet \cite{Xu_2022_CVPR}, the graph-based SSAGCN \cite{lv2023ssagcn}, and the transformer-based AutoBot \cite{girgis2022latent}. We also conducted comprehensive experiments using three trajectory prediction benchmark datasets: the ETH-UCY \cite{pellegrini2009you, lerner2007crowds} dataset, the NBA SportVU dataset \cite{zhan2018generating}, and the Trajnet++ \cite{Kothari2020HumanTF} dataset. The consistently superior results validate the efficacy of our approach.

The key contributions of our work are threefold. 
\begin{itemize}
    \item First, while trajectory prediction methods typically implicitly learn temporal representations, we introduce a novel approach called  \textit{TWD} that explicitly incorporates temporal dependencies during the training of a trajectory prediction model. This is achieved by stochastically dropping waypoints from past observed trajectories. The model is compelled to learn the underlying temporal representation from the remaining waypoints, resulting in an enhanced understanding of temporal patterns.
    \item Second, our approach ensures that the model is well-suited for scenarios where observed waypoint sequences may be missing or incomplete. The inclusion of stochastic temporal waypoint dropping in the model learning process significantly improves its performance in situations with missing or incomplete values.
    \item Third, empirical results demonstrate that our approach substantially enhances the model's capabilities in trajectory prediction. Our approach can be a valuable complement to existing trajectory prediction methods, improving their prediction accuracy.
\end{itemize}

\section{Related Work}

\subsection{Trajectory Prediction} Earlier studies in trajectory prediction primarily concentrated on deterministic methodologies, which involved investigating force models \cite{helbing1995social}, employing recurrent neural networks (RNNs) \cite{salzmann2020trajectron++}, and employing frequency analysis \cite{park2020diverse} techniques. Force models \cite{helbing1995social} in trajectory prediction use a measure to represent the internal motivations of individuals for performing specific actions. These models typically incorporate variables such as acceleration and desired velocity to model the trajectories of objects or agents. Approach proposed in \cite{park2020diverse} is capable of modeling multiple plausible sequences of actions that agents can take to reach their intended goals while adhering to physical constraints and staying within drivable areas. Trajectron++ \cite{salzmann2020trajectron++} is a modular graph-structured recurrent model designed to forecast the trajectories of a diverse group of agents. 

As predicting the future trajectory of an agent involves inherent uncertainty and often results in multiple possible outcomes, recent advancements in trajectory prediction methods have adopted the utilization of deep generative models \cite{shi2021sgcn,gupta2018social,lee2022muse,xu2022dynamic,mangalam2020not}. These models encompass various techniques, including conditional variational autoencoders (CVAEs) \cite{Xu_2022_CVPR,lee2022muse,xu2022dynamic,mangalam2020not}, generative adversarial networks (GANs) \cite{sophie19, hu2020collaborative,gupta2018social,kosaraju2019social}, and diffusion models \cite{mao2023leapfrog}. CVAE is employed to estimate the parameters of a latent distribution, allowing for the sampling of future trajectory features from latent distribution. Groupnet \cite{Xu_2022_CVPR} excels at capturing interactions and the foundational representation needed for predicting socially plausible trajectories through relational reasoning. DyngroupNet \cite{xu2022dynamic} possesses the ability to capture time-varying interactions, encompassing both pair-wise and group-wise interactions. PecNet \cite{mangalam2020not} and Muse \cite{lee2022muse} models are designed for long-range multi-modal trajectory prediction, allowing them to infer a wide range of trajectory possibilities while ensuring social compliance among agents. Nmmp \cite{hu2020collaborative} explicitly incorporates interaction modeling and the learning of representations for directed interactions between actors. Sophie \cite{sophie19} utilizes social attention mechanisms to incorporate both physical and social information. Sgan \cite{gupta2018social} and Sophie \cite{sophie19} leverage GANs to produce more realistic trajectory samples and capture the inherent uncertainty in future paths. The method proposed by \cite{shi2023representing} addresses the issue of superfluous interactions. It introduces the Interpretable Multimodality Predictor (IMP), describing the mean location distribution as a Gaussian Mixture Model (GMM). This method encourages multimodality by sampling various mean waypoints of trajectories. Emerging diffusion models \cite{gu2022stochastic,mao2023leapfrog} have demonstrated their significant representation capabilities in stochastic trajectory prediction. However, they suffer from high inference times due to large denoising steps. LED \cite{mao2023leapfrog} addresses this limitation by introducing a trainable leapfrog initializer.

The application of the Transformer architecture \cite{vaswani2017attention} in this domain \cite{girgis2022latent,yuan2021agentformer,zhou2023query,yu2020spatio} has gained attention because of its modeling of spatio-temporal relations through attention mechanisms. They include attention layers for encoding sequential features in the temporal dimension, self-attention layers for capturing interactions among traffic participants in the social dimension, and cross-attention layers dedicated to learning agents patterns. In addition, significant advancements in goal-based methods \cite{
chiara2022goal,bae2023set} have produced good outcomes. These methods initially predict agents' intentions based on endpoints of trajectories, followed by trajectory prediction refinement. Various studies \cite{girgis2022latent,sekhon2021scan} have also produced individual predictions for each agent separately, while others have ventured into generating concurrent joint predictions. SRGAT \cite{chen2023goal} combines multiple anticipated goals for each agent, followed by social interaction modeling. Furthermore, Graph-Based approaches \cite{lv2023ssagcn,sekhon2021scan,pmlr-v80-kipf18a} have emerged as means to model agent interactions, especially within non-grid structures. SSAGCN \cite{lv2023ssagcn} employed a graph neural network to capture social and scene interactions. Additionally, a social soft attention function enabled quantifying the influence degree among pedestrians. DynGroupNet \cite{xu2023dynamic} and TDGCN \cite{wang2023trajectory} focus on temporal groupwise interactions, considering interaction strength and category for modeling future predictions. 
For integrating environmental information, some methods  \cite{shi2021sgcn,mendieta2021carpe} involve the encoding of scene information using convolutional neural networks. DISTLR \cite{cao2023discovering} formulates a set of spatial-temporal logic sets to characterize human actions. Furthermore, MERA \cite{sun2023modality} employs different modalities in motion predictions, processing various feature clusters to represent modalities such as scene semantics and agent motion state. The Multi-Style Network (MSN) \cite{wong2023msn} integrates style as an element in predictions, enabling multi-style predictions for trajectories. MetaTraj \cite{shi2023metatraj} provides trajectory prediction sub-tasks and a meta-task capable of handling predictions for unobserved scenes and objects.

\subsection{Regularizing Trajectory Prediction}
Several techniques have been explored in recent studies to enhance the quality of representation learning for the accurate prediction of future trajectories. Some approaches \cite{ye2022bootstrap} involve applying diverse transformations to the same input data to derive perturbation-invariant representations. Alternatively, some researchers \cite{ girgis2022latent,sekhon2021scan,zhu2020robust} reverse the temporal order of trajectories and establish pairwise consistency between the resulting predictions. Another strategy introduced in prior work \cite{aydemir2023adapt} centers on achieving consistency by scrutinizing the disparities between agent-centric and scene-centric settings. Some methods \cite{ye2022bootstrap} incorporate a spatial permutation function encompassing operations like flipping and introducing random noise applied to trajectories from the initial stage. It is posited that, even under minor spatial permutations and disturbances, the network's outputs should maintain self-consistency. Different from traditional data augmentation, these methodologies can be described as explicit regularization. 

\subsection{Temporal Learning}
In terms of modeling temporal motion patterns of trajectories, various architectural designs have been introduced for trajectory prediction like Social-LSTM \cite{alahi2016social} and STGAT \cite{huang2019stgat}. STS-LSTM \cite{zhang2023spatial} employs LSTM to effectively modle spatiotemporal interactions. Similarly Models like Social Attention \cite{vemula2018social} and Trajectron \cite{ivanovic2019trajectron} use LSTM in order to generate a spatiotemporal graph that is capable of representing structured sequence data for obtaining optimal outcomes. RNN \cite{salzmann2020trajectron++,park2020diverse} based models are also predominant in learning temporal flow, but they may suffer from vanishing gradient issues in some situations. Some models, such as SGCN \cite{shi2021sgcn}, use temporal convolutional networks (TCN) \cite{lv2023ssagcn} to learn temporal representation. Transformer \cite{vaswani2017attention} is capable of understanding long-term dependencies in a better way with an effective self-attention method. In trajectory prediction, transformers also perform very well in accurately predicting the trajectories.  Recent works \cite{girgis2022latent,yuan2021agentformer,zhou2023query} have utilized transformers for simulating temporal dependence and enhancing performance. 
AutoBot \cite{girgis2022latent} is an encoder-decoder framework that creates consistent trajectories for multiple agents in a scene. They utilize a latent variable sequential set transformer. AgentFormer \cite{yuan2021agentformer} can model both the temporal and social aspects. This model works by representing multi-agent trajectories as a sequence, combining trajectory features across time and agents. VIKT \cite{zhong2023visual} incorporates visual localization and orientation to enhance trajectory prediction by learning from real-world visual settings. They also leverage Visual Intention Knowledge (VIK) with the spatiotemporal Transformer (VIKT) to represent human intent. VNAGT \cite{chen2023vnagt} introduces a variational non-autoregressive graph transformer to capture social and temporal interactions. LSSTA \cite{yang2023long} proposes a spatial transformer that effectively models the dynamic nature of pedestrian interactions while also accounting for time-varying spatial dependencies. There exists the accumulation error. To mitigate the accumulation of prediction errors, SIM \cite{li2023synchronous} introduces a synchronous bi-directional structure. Query-based method \cite{zhou2023query} employs query-centric techniques and anchor-free queries to generate trajectory proposals in a recurrent manner. 

Our approach complements existing methods by learning temporal representations under constraints, such as incomplete temporal sequences. This can effectively mitigate biases that might lead the model to rely excessively on specific temporal sequences when making predictions. Our approach explores temporal waypoint dropping and demonstrates its effectiveness in enhancing existing approaches.

\begin{figure*}[t]
    \centering
    \includegraphics[scale=.24]{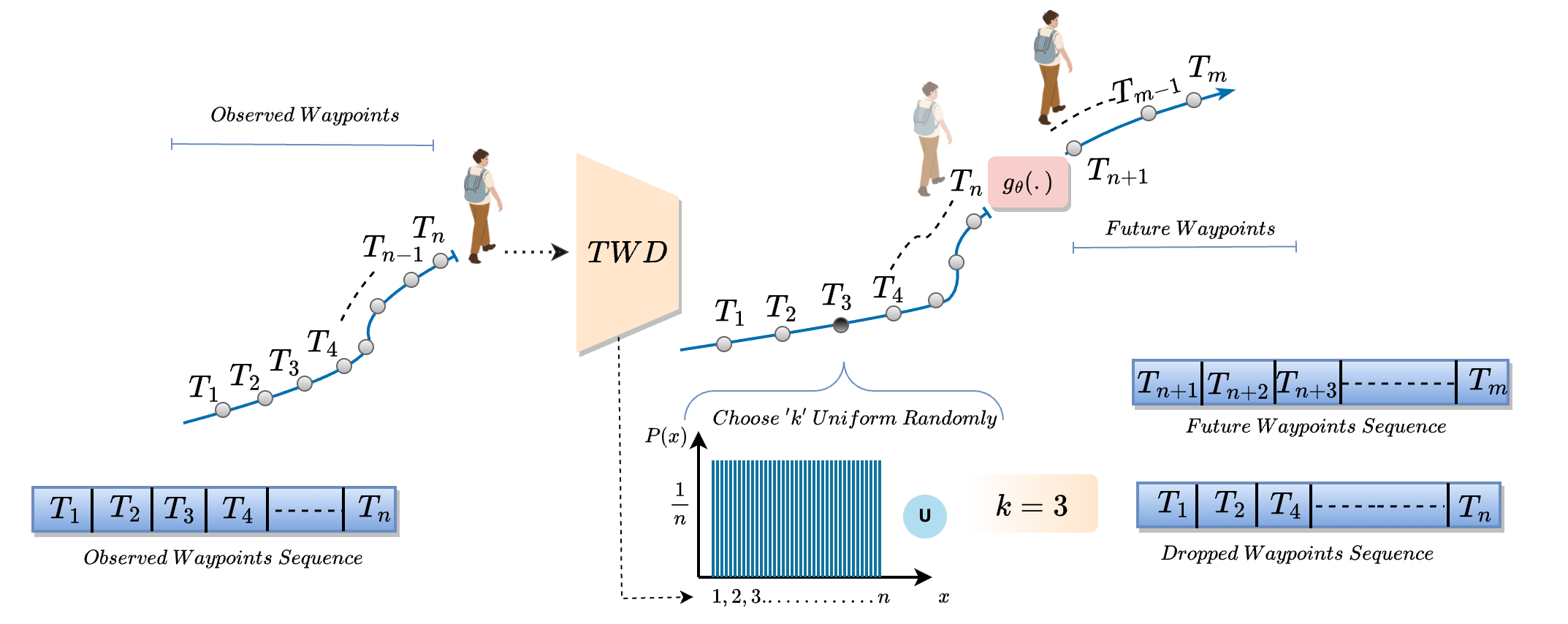}
    \caption{Visualizing the stochastic temporal waypoint dropping process, where the observed waypoint sequence ($T_1, \dots, T_n$) is input into the TWD (Temporal Waypoint Dropping). A waypoint to be dropped is uniformly and randomly chosen (depicted in black at time stamp $T_k$, with $k=3$) from the past observed temporal sequence. The value of $k$ is sampled from a uniform distribution, and the corresponding waypoint at time stamp $T_k$ is then removed from the past observed trajectory. Finally, the modified past observed temporal sequence after waypoint dropping is used as input to train a prediction model $g_{\theta}(\cdot)$ for forecasting the future trajectory ($T_{n+1}, \dots, T_m$).}
    \label{scot}
\end{figure*}

\section{Methodology} \label{method}

\subsection{Problem Definition}

let $\mathbf{X_{i}} = [\mathbf{x_{i}}^{T_1}, \dots, \mathbf{x_{i}}^{T_{\rm n -1}}, \mathbf{x_{i}}^{T_{\rm n}}]\in \mathbb{R}^{T_{\rm n} \times 2}$ be the trajectory observed in the past timestamps for $i^{th}$ agent, where $\mathbf{x_{i}}^{T_n}\in \mathbb{R}^2$ is the coordinate $(x,y)$ of the $n^{th}$ waypoint at $T_n$ observed time. $|X_i|=n$ represents the number of waypoints in the $X_i$ sequence. The main objective of this problem setting is to train a prediction model $g_{\theta}(\cdot)$ with parameters $\theta$ to produce future trajectories labeled as $P_{\theta}$ = $g_{\theta}(X_i)$. We sample predicted future trajectory $\widehat{\mathbf{Y_{i}}} = [\mathbf{y_{i}}^{T_{n+1}}, \mathbf{y_{i}}^{T_{n+2}}, \dots, \mathbf{y_{i}}^{T_{\rm m}}]\in \mathbb{R}^{T_{\rm m} \times 2}$, which should be closely aligns with the actual future trajectory $\mathbf{Y_{i}}$ (ground truth). Here ${y_{i}}^{T_{m}}\in \mathbb{R}^{2}$ is the coordinate of $m^{th}$ waypoint at predicted future time instance $T_{m}$. Also, $|Y_i|=m$ represents the number of waypoints in the future trajectory. The overall problem formulation for trajectory prediction is given below, where $DEV$ measures deviation from ground truth trajectory.

\begin{equation}
\theta^*=\min _\theta \min _{\widehat{\mathbf{Y}} \in \hat{\mathcal{Y}}} DEV\left(\widehat{\mathbf{Y_{i}}}, \mathbf{Y_{i}}\right), \text { s.t. } \widehat{\mathcal{Y}} \sim \mathcal{P}_\theta .
\end{equation}

\subsection{Single Temporal Waypoint Dropping}
\label{stwd}
In Temporal Waypoint Dropping (\textit{TWD}), we sample from the past trajectories $\omega = \{{X_1, X_2, X_3, \dots, X_{N_{}^{}}}\}$ of all $N$ agents, where $X_i$ is the $i^{th}$ agent's past trajectory. We then generate a sequence where a temporal waypoint is dropped for all agents (i.e., $TWD(\omega)$). We employ a stochastic drop strategy to drop waypoints from past observed trajectories. Stochastic drop introduces randomness in dropping. Finally, model $g_{\theta}(\cdot)$ predicts the future trajectories $\tau = \{{\widehat{Y_1}, \widehat{Y_2}, \widehat{Y_3}, \dots, \widehat{Y_{N_{}^{}}}}\}$ for all agents given $TWD(\omega)$. The final objective is finding the parameter $\theta$ that maximizes the expected log-likelihood of future trajectories given the observed data $TWD(\omega)$.

\begin{equation}
\max _\theta \mathbb{E}_{\boldsymbol{\omega}} \sum_{n=1}^N \sum_{t=T_{n+1}}^{T_m} \log P_\theta\left(\mathbf{Y}_n^t \mid \operatorname{TWD}(\boldsymbol{\omega})\right)
\end{equation}

\begin{equation}
\label{eq_twdinput}
TWD(\omega) = TWD\{{X_1, X_2, X_3, \dots, X_{N_{}^{}}}\}
\end{equation}

Here, $T_m$ represents the future timestamp, and $N$ denotes the number of agents. Please note that \textit{TWD} in Equation \ref{eq_twdinput} takes observed trajectories as input.

\begin{gather}
\label{eq_twdinput2}
TWD(\omega) = TWD\{ (x_{1}^{T_1}, x_{1}^{T_2}, x_{1}^{T_3},\dots, x_{1}^{T_n}), (x_{2}^{T_1}, x_{2}^{T_2},\notag\\ x_{2}^{T_3}, \dots, x_{2}^{T_n})
,\dots, (x_{N}^{T_1}, x_{N}^{T_2}, x_{N}^{T_3},\dots, x_{N}^{T_n})\}
\end{gather}

Given the past observed trajectories, we stochastically drop the temporal waypoint from the given sequence during the training of a trajectory prediction model. This ensures that the model performs better, even in scenarios where observed waypoint sequences may be missing or incomplete.

\begin{equation} 
\label{eqfour}
    TWD(\omega) = \{S_1^k, S_2^k, S_3^k, \ldots, S_N^k\}
\end{equation} 

In Equation \ref{eqfour}, the trajectory sequence $S_{i}^{k}$ includes all waypoints of $X_i$ except the $k^{th}$ time stamp waypoint (i.e., $S_{i}^{k} = [\mathbf{x_{i}}^{T_1}, \dots, \mathbf{x_{i}}^{T_{\rm k-1}}, \mathbf{x_{i}}^{T_{\rm k+1}}, \dots, \mathbf{x_{i}}^{T_{\rm n}}]$ ). In $S_{i}^{k}$, $k^{th}$ time stamp waypoint is dropped from $X_i$ of $i^{th}$ agent and the selection of $k$ can be done using stochastic processes given below.

\paragraph{Stochastic Process} In the stochastic dropping process, we assume a uniform distribution for waypoint removal, as each waypoint is equally likely to be removed (see Figure~\ref{scot}). This ensures fairness in the dropping process through the random removal of waypoints from the trajectories. The selection of $k$ is determined by the Equation below.
\begin{equation}
\label{eqfive}
   k =Uniform Random\{1, 2, 3, \dots,  n\}
\end{equation}

Here, $k$ is a value that can be uniformly and randomly chosen from the set $\{1, 2, 3, \dots, n\}$ with a probability of $\frac{1}{n}$. The corresponding waypoint at time stamp $T_k$ is then dropped from the past trajectories of all $N$ agents (see Equation~\ref{eqfour}).

\subsection{Multiple Temporal Waypoints Dropping}
\label{s_mtwd}
We discussed the single temporal waypoint dropping process in Section~\ref{stwd}. We can iterate the process given in Section~\ref{stwd} multiple times to drop multiple temporal waypoints. The objective function in multiple temporal waypoints dropping is given below.

\begin{equation}
\max _\theta \mathbb{E}_{\boldsymbol{\omega}} \sum_{n=1}^N \sum_{t=T_{n+1}}^{T_m} \log P_\theta\left(\mathbf{Y}_n^t \mid \operatorname{TWD^D}(\boldsymbol{\omega})\right)
\end{equation}

\begin{align}
\label{eq_mul_twd}
    TWD^D(\omega) = \underbrace{\text{TWD}(\text{TWD}(\ldots(\text{TWD}(w))))}_{D \text{ times}}
\end{align}

Here, $D$ represents the number of waypoint drops from past observed temporal sequences. For instance, to drop two waypoints (i.e., $D=2$) from the past trajectories of all agents, we will iterate the \textit{TWD} two times (i.e., $\text{TWD}(\text{TWD}(w))$). In each iteration, the stochastic dropping process selects the value of $k$ uniformly and randomly from the currently available set of past observed temporal sequences for dropping. Subsequently, the waypoint corresponding to the timestamp $T_k$ is dropped from the past trajectories of all $N$ agents, as mentioned in Section~\ref{stwd}. Please note that $D$ is a hyperparameter and can be determined using cross-validation. We conducted ablation experiments (see Section~\ref{abalation_multodrop} and Table~\ref{abs_mult}) to demonstrate the impact of single vs. multiple temporal waypoint drops on the model’s predictive performance.

\subsection{Training and Evaluation}

During training, the model $g_{\theta}(.)$ uses trajectory sequences generated by either $TWD(\omega)$ or $TWD^D(\omega)$ as input, depending on whether a single temporal waypoint is dropped or multiple temporal waypoints are dropped. The modified past observed temporal sequences after waypoint dropping are generated using the process described in Section~\ref{stwd} (for $TWD(\omega)$) and Section~\ref{s_mtwd} (for $TWD^D(\omega)$), which are then used as input in the training of the model $g_{\theta}(.)$. Please note that in each iteration of training, different temporal waypoints are dropped. We conducted ablation experiments (see Section~\ref{ablation}) to provide insights into the impact of these processes on the model's prediction performance. The evaluation of the model $g_{\theta}(.)$ performance is calculated on the standard test split across the standard datasets with default settings. More details are provided in the experimentation Section~\ref{section_Experimen}.

\section{Experiments} \label{section_Experimen}

In this Section, we present the quantitative and qualitative results of our approach (\textit{TWD}). Additionally, we conduct ablation studies to assess the ability of \textit{TWD} to improve trajectory prediction. 

\subsection{Experimental Setup}

\subsubsection{Datasets} We evaluate the performance of \textit{TWD} on three benchmarked trajectory datasets: NBA \cite{zhan2018generating}, TrajNet++ \cite{Kothari2020HumanTF}, and ETH-UCY \cite{pellegrini2009you,lerner2007crowds}. The NBA sports VU dataset encompasses trajectory data for all ten players during live NBA games, where both teammates from teams and the basketball court layout significantly impact player movements. We follow the setting, which involves forecasting the subsequent ten timestamps, which is equivalent to 4.0 seconds, based on the preceding five timestamps spanning 2.0 seconds of historical timestamps. TrajNet++ is another dataset focused on agent-agent interactions, emphasizing substantial interaction among agents within a scene. We chose TrajNet++ due to its deliberate design aimed at facilitating high levels of interaction among scene agents, as described by Kothari et al. \cite{Kothari2020HumanTF}. Our model evaluation is conducted on the synthetic segment of TrajNet++, where we predict the 12 future timestamps based on the past 9 timestamps of the agents.
 ETH-UCY is a mixture of two datasets with smooth trajectories and simple agent interactions. The ETH dataset comprises two scenes, ETH and HOTEL, with 750 pedestrians. The UCY contains three scenes with 786 pedestrians, including UNIV, ZARA1, and ZARA2. The scenes contain the road, a crossroads, and a nearly open space. We are provided with  a world-coordinate sequence, which consists of a trajectory spanning 8 time steps, equivalent to 3.2 seconds. Our objective is to forecast the next 12 time steps, thereby extending our predictions to cover a total duration of 4.8 seconds.

\subsubsection{Evaluation Metric} 
Our experiments employ standard evaluation metrics for trajectory prediction, including Average Displacement Error (ADE) and Final Displacement Error (FDE). ADE represents the average L2 distance between predicted and ground truth trajectories across all time steps, while FDE quantifies the L2 distance at the last time step or final endpoint.

\subsubsection{Implementation Details}  
For a fair comparison with the compared methods, we retained their default settings, including the sequence length of trajectories and timestamps used as model input. To maintain the original temporal sequence length, we repeated the initial waypoints. It is important to note that during training, we employed single temporal waypoint dropping in all our experiments. 

\subsubsection{Baseline Models} 
We evaluate \textit{TWD} by assessing it on three distinct models, each representing a different framework: Variational Autoencoder-based (GroupNet \cite{Xu_2022_CVPR}), Graph-based (SSAGCN \cite{lv2023ssagcn}), and Transformer-based (AutoBot \cite{girgis2022latent}). The SSAGCN model effectively handles the pedestrian's social and scene interaction and predicts the trajectories that are both socially and physically plausible. They use the spatial-temporal graph to model the degree of influence among pedestrians and scene attention to model the physical effects of the environment. GroupNet can capture interactions and the underlying representation for predicting socially plausible trajectories with relational reasoning. GroupNet with conditional variational autoencoder can model complex social influences for better trajectory prediction. AutoBot is an encoder-decoder framework that constructs scene-consistent multi-agent trajectories based on latent variable sequential set transformer. The encoder comprises alternating temporal and social multi-head self-attention mechanisms that perform equivariant computations over time and social dimensions.

\begin{table}[]
\caption{The minimum Average Displacement Error (minADE) and minimum Final Displacement Error (minFDE) for prediction on the NBA dataset using the \textit{TWD} method. (B) denotes the baseline model. RD($\%$) indicates the relative percentage difference compared to the baseline.}
\label{nba_dataset}
\resizebox{\columnwidth}{!}{%
\begin{tabular}{c|cccc}
\hline
\multirow{2}{*}{\textbf{Method}} &
  \multicolumn{4}{c}{\textbf{Time}} \\ \cline{2-5} 
 &
  1.0s &
  2.0s &
  3.0s &
  4.0s \\ \hline
\begin{tabular}[c]{@{}c@{}}S-LSTM   \cite{alahi2016social}\end{tabular} &
  \multicolumn{1}{c|}{0.45/0.67} &
  \multicolumn{1}{c|}{0.88/1.53} &
  \multicolumn{1}{c|}{1.33/2.38} &
  1.79/3.16 \\ \hline
\begin{tabular}[c]{@{}c@{}}Social-GAN  \cite{gupta2018social}\end{tabular} &
  \multicolumn{1}{c|}{0.46/0.65} &
  \multicolumn{1}{c|}{0.85/1.36} &
  \multicolumn{1}{c|}{1.24/1.98} &
  1.62/2.51 \\ \hline
\begin{tabular}[c]{@{}c@{}}Social-STGCNN \cite{mohamed2020social}\end{tabular} &
  \multicolumn{1}{c|}{0.36/0.50} &
  \multicolumn{1}{c|}{0.75/0.99} &
  \multicolumn{1}{c|}{1.15/1.79} &
  1.59/2.37 \\ \hline
\begin{tabular}[c]{@{}c@{}}STGAT \cite{huang2019stgat}\end{tabular} &
  \multicolumn{1}{c|}{0.38/0.55} &
  \multicolumn{1}{c|}{0.73/1.18} &
  \multicolumn{1}{c|}{1.07/1.74} &
  1.41/2.22 \\ \hline
\begin{tabular}[c]{@{}c@{}}NRI \cite{pmlr-v80-kipf18a}\end{tabular} &
  \multicolumn{1}{c|}{0.45/0.64} &
  \multicolumn{1}{c|}{0.84/1.44} &
  \multicolumn{1}{c|}{1.24/2.18} &
  1.62/2.84 \\ \hline
\begin{tabular}[c]{@{}c@{}}STAR \cite{yu2020spatio}\end{tabular} &
  \multicolumn{1}{c|}{0.43/0.65} &
  \multicolumn{1}{c|}{0.77/1.28} &
  \multicolumn{1}{c|}{1.00/1.55} &
  1.26/2.04 \\ \hline
\begin{tabular}[c]{@{}c@{}}PECNet \cite {mangalam2020not}\end{tabular} &
  \multicolumn{1}{c|}{0.51/0.76} &
  \multicolumn{1}{c|}{0.96/1.69} &
  \multicolumn{1}{c|}{1.41/2.52} &
  1.83/3.41 \\ \hline
\begin{tabular}[c]{@{}c@{}}NMMP \cite{hu2020collaborative}\end{tabular} &
  \multicolumn{1}{c|}{0.38/0.54} &
  \multicolumn{1}{c|}{0.70/1.11} &
  \multicolumn{1}{c|}{1.01/1.61} &
  1.33/2.05 \\ \hline
\begin{tabular}[c]{@{}c@{}}Stimulus Verification \cite{sun2023stimulus}\end{tabular} &
  \multicolumn{1}{c|}{-} &
  \multicolumn{1}{c|}{-} &
  \multicolumn{1}{c|}{-} &
  1.08/1.12 \\ \hline
\begin{tabular}[c]{@{}c@{}}TDGCN \cite{wang2023trajectory}\end{tabular} &
  \multicolumn{1}{c|}{0.30/0.45} &
  \multicolumn{1}{c|}{0.53/0.82} &
  \multicolumn{1}{c|}{0.80/117} &
  1.06/153 \\ \hline
\begin{tabular}[c]{@{}c@{}}MERA \cite{sun2023modality}\end{tabular} &
  \multicolumn{1}{c|}{-} &
  \multicolumn{1}{c|}{-} &
  \multicolumn{1}{c|}{-} &
  1.17/2.21 \\ \hline
\begin{tabular}[c]{@{}c@{}}DISTL  \cite{cao2023discovering}\end{tabular} &
  \multicolumn{1}{c|}{0.30/0.40} &
  \multicolumn{1}{c|}{0.58/0.88} &
  \multicolumn{1}{c|}{0.87/1.31} &
  1.13/1.60 \\ \hline
\begin{tabular}[c]{@{}c@{}}GroupNet (B) \cite{Xu_2022_CVPR}\end{tabular} &
  \multicolumn{1}{c|}{0.34/0.48} &
  \multicolumn{1}{c|}{0.62/0.95} &
  \multicolumn{1}{c|}{0.87/1.31} &
  1.13/1.69 \\ \hline
\begin{tabular}[c]{@{}c@{}}Our (B + TWD)\end{tabular} &
  \multicolumn{1}{c|}{\textbf{0.24/0.32}} &
  \multicolumn{1}{c|}{\textbf{0.46/0.65}} &
  \multicolumn{1}{c|}{\textbf{0.69/0.95}} &
  \textbf{0.92/1.19}  \\ \hline
\begin{tabular}[c]{@{}c@{}}RD($\%$) ADE/FDE\end{tabular} &
  \multicolumn{1}{l|}{\textbf{34.5/40.0}} &
  \multicolumn{1}{l|}{\textbf{29.6/37.5}} &
  \multicolumn{1}{l|}{\textbf{23.1/31.8}} &
  \multicolumn{1}{l}{\textbf{20.5/34.7}} \\ \hline
\end{tabular}%
}
\end{table}

\begin{table*}[]
\centering
\caption{Minimum ADE ($\downarrow$) / Minimum FDE ($\downarrow$) for trajectory prediction on the ETH-UCY dataset utilizing the \textit{TWD} technique during training. (B) denotes the baseline model. RD($\%$) indicates the relative percentage difference  compared to the baseline.}
\label{table_eth}
\begin{tabular}{|c|ccccc|c|}
\hline
\textbf{Method} &
  \textbf{ETH} &
  \textbf{HOTEL} &
  \textbf{UNIV} &
  \textbf{ZARA1} &
  \textbf{ZARA2} &
  \textbf{AVG} \\ \hline
\begin{tabular}[c]{@{}c@{}}SGAN \cite{gupta2018social} \end{tabular} &
  \multicolumn{1}{c|}{0.87/1.62} &
  \multicolumn{1}{c|}{0.67/1.37} &
  \multicolumn{1}{c|}{0.76/1.52} &
  \multicolumn{1}{c|}{0.35/0.68} &
  0.42/0.84 &
  0.61/1.21 \\ \hline
\begin{tabular}[c]{@{}c@{}}Sophie \cite{sophie19} \end{tabular} &
  \multicolumn{1}{c|}{0.70/1.43} &
  \multicolumn{1}{c|}{0.76/1.67} &
  \multicolumn{1}{c|}{0.54/1.24} &
  \multicolumn{1}{c|}{0.30/0.63} &
  0.38/0.78 &
  0.54/1.15 \\ \hline
\begin{tabular}[c]{@{}c@{}}STGAT \cite{sekhon2021scan} \end{tabular} &
  \multicolumn{1}{c|}{0.56/1.10} &
  \multicolumn{1}{c|}{0.27/0.50} &
  \multicolumn{1}{c|}{0.32/0.66} &
  \multicolumn{1}{c|}{0.21/0.42} &
  0.20/0.40 &
  0.31/0.62 \\ \hline
\begin{tabular}[c]{@{}c@{}}Social-BiGAT \cite{kosaraju2019social} \end{tabular} &
  \multicolumn{1}{c|}{0.69/1.29} &
  \multicolumn{1}{c|}{0.49/1.01} &
  \multicolumn{1}{c|}{0.55/1.32} &
  \multicolumn{1}{c|}{0.30/0.62} &
  0.36/0.75 &
  0.48/1.00 \\ \hline
\begin{tabular}[c]{@{}c@{}}NMMP \cite{hu2020collaborative} \end{tabular} &
  \multicolumn{1}{c|}{0.62/1.08} &
  \multicolumn{1}{c|}{0.33/0.63} &
  \multicolumn{1}{c|}{0.52/1.11} &
  \multicolumn{1}{c|}{0.32/0.66} &
  0.29/0.61 &
  0.41/0.82 \\ \hline
\begin{tabular}[c]{@{}c@{}}Social-STGCNN \cite{mohamed2020social} \end{tabular} &
  \multicolumn{1}{c|}{0.64/1.11} &
  \multicolumn{1}{c|}{0.49/0.85} &
  \multicolumn{1}{c|}{0.44/0.79} &
  \multicolumn{1}{c|}{0.34/0.53} &
  0.30/0.48 &
  0.44/0.75 \\ \hline
\begin{tabular}[c]{@{}c@{}}CARPE \cite{mendieta2021carpe} \end{tabular} &
  \multicolumn{1}{c|}{0.80/1.4} &
  \multicolumn{1}{c|}{0.52/1.00} &
  \multicolumn{1}{c|}{0.61/1.23} &
  \multicolumn{1}{c|}{0.42/0.84} &
  0.34/0.74 &
  0.46/0.89 \\ \hline
\begin{tabular}[c]{@{}c@{}}PecNet \cite {mangalam2020not} \end{tabular} &
  \multicolumn{1}{c|}{0.54/0.87} &
  \multicolumn{1}{c|}{0.18/0.24} &
  \multicolumn{1}{c|}{0.35/0.60} &
  \multicolumn{1}{c|}{0.22/0.39} &
  0.17/0.30 &
  0.29/0.48 \\ \hline
\begin{tabular}[c]{@{}c@{}}Trajectron++ \cite{salzmann2020trajectron++} \end{tabular} &
  \multicolumn{1}{c|}{0.43/0.86} &
  \multicolumn{1}{c|}{0.12/0.19} &
  \multicolumn{1}{c|}{0.22/0.43} &
  \multicolumn{1}{c|}{0.17/0.32} &
  0.12/0.25 &
  0.21/0.41 \\ \hline
\begin{tabular}[c]{@{}c@{}}GTPPO \cite{yang2021novel} \end{tabular} &
  \multicolumn{1}{c|}{0.63/0.98} &
  \multicolumn{1}{c|}{0.19/0.30} &
  \multicolumn{1}{c|}{0.35/0.60} &
  \multicolumn{1}{c|}{0.20/0.32} &
  0.18/0.31 &
  0.31/0.50 \\ \hline
\begin{tabular}[c]{@{}c@{}}SGCN \cite{shi2021sgcn} \end{tabular} &
  \multicolumn{1}{c|}{0.52/1.03} &
  \multicolumn{1}{c|}{0.32/0.55} &
  \multicolumn{1}{c|}{0.37/0.70} &
  \multicolumn{1}{c|}{0.29/0.53} &
  0.25/0.45 &
  0.37/0.65 \\ \hline
\begin{tabular}[c]{@{}c@{}}Introvert \cite{shafiee2021introvert} \end{tabular} &
  \multicolumn{1}{c|}{0.42/0.70} &
  \multicolumn{1}{c|}{0.11/0.17} &
  \multicolumn{1}{c|}{0.20/0.32} &
  \multicolumn{1}{c|}{0.16/0.27} &
  0.16/0.25 &
  0.21/0.34 \\ \hline
\begin{tabular}[c]{@{}c@{}}LB-EBM \cite{pang2021trajectory} \end{tabular} &
  \multicolumn{1}{c|}{0.30/0.52} &
  \multicolumn{1}{c|}{0.13/0.20} &
  \multicolumn{1}{c|}{0.27/0.52} &
  \multicolumn{1}{c|}{0.20/0.37} &
  0.15/0.29 &
  0.21/0.38 \\ \hline
\begin{tabular}[c]{@{}c@{}}Y-Net \cite{mangalam2021goals} \end{tabular} &
  \multicolumn{1}{c|}{0.28/0.33} &
  \multicolumn{1}{c|}{0.10/0.14} &
  \multicolumn{1}{c|}{0.24/0.41} &
  \multicolumn{1}{c|}{0.17/0.27} &
  0.13/0.22 &
  0.18/0.27 \\ \hline
\begin{tabular}[c]{@{}c@{}}SSAGCN (B) \cite{lv2023ssagcn} \end{tabular} &
  \multicolumn{1}{c|}{0.21/0.38} &
  \multicolumn{1}{c|}{0.11/0.19} &
  \multicolumn{1}{c|}{0.14/0.25} &
  \multicolumn{1}{c|}{0.12/0.22} &
  0.09/0.15 &
  0.13/0.24 \\ \hline
\begin{tabular}[c]{@{}c@{}}GroupNet  \cite{Xu_2022_CVPR}\end{tabular} &
  \multicolumn{1}{c|}{0.46/0.73} &
  \multicolumn{1}{c|}{0.15/0.25} &
  \multicolumn{1}{c|}{0.26/0.49} &
  \multicolumn{1}{c|}{0.21/0.39} &
  \multicolumn{1}{c|}{0.17/0.33} &
  0.25/0.44 \\ \hline
\begin{tabular}[c]{@{}c@{}}DynGroupNet \cite{xu2023dynamic}\end{tabular} &
  \multicolumn{1}{c|}{0.42/0.66} &
  \multicolumn{1}{c|}{0.13/0.20} &
  \multicolumn{1}{c|}{0.24/0.44} &
  \multicolumn{1}{c|}{0.19/0.34} &
  \multicolumn{1}{c|}{0.15/0.28} &
  0.23/0.38 \\ \hline
\begin{tabular}[c]{@{}c@{}}TDGCN \cite{wang2023trajectory}\end{tabular} &
  \multicolumn{1}{c|}{0.51/0.68} &
  \multicolumn{1}{c|}{0.25/0.44} &
  \multicolumn{1}{c|}{0.30/0.50} &
  \multicolumn{1}{c|}{0.24/0.42} &
  \multicolumn{1}{c|}{0.16/0.27} &
  0.29/0.46 \\ \hline

\begin{tabular}[c]{@{}c@{}}MERA \cite{sun2023modality}\end{tabular} &
  \multicolumn{1}{c|}{0.26/0.50} &
  \multicolumn{1}{c|}{0.11/0.19} &
  \multicolumn{1}{c|}{0.25/0.53} &
  \multicolumn{1}{c|}{0.19/0.40} &
  \multicolumn{1}{c|}{0.15/0.31} &
  0.19/0.39 \\ \hline
\begin{tabular}[c]{@{}c@{}}RMB \cite{shi2023representing}\end{tabular} &
  \multicolumn{1}{c|}{0.29/0.49} &
  \multicolumn{1}{c|}{0.12/0.18} &
  \multicolumn{1}{c|}{0.29/0.51} &
  \multicolumn{1}{c|}{0.20/0.36} &
  \multicolumn{1}{c|}{0.15/0.27} &
  0.21/0.36 \\ \hline
\begin{tabular}[c]{@{}c@{}}VIKT  \cite{zhong2023visual}\end{tabular} &
  \multicolumn{1}{c|}{0.30/0.51} &
  \multicolumn{1}{c|}{0.13/0.25} &
  \multicolumn{1}{c|}{0.23/0.51} &
  \multicolumn{1}{c|}{0.21/0.44} &
  \multicolumn{1}{c|}{0.14/0.30} &
  0.20/0.40 \\ \hline

\begin{tabular}[c]{@{}c@{}}MSN \cite{wong2023msn}\end{tabular} &
  \multicolumn{1}{c|}{0.27/0.41} &
  \multicolumn{1}{c|}{0.11/0.17} &
  \multicolumn{1}{c|}{0.28/0.48} &
  \multicolumn{1}{c|}{0.22/0.36} &
  \multicolumn{1}{c|}{0.18/0.29} &
  0.21/0.34 \\ \hline
\begin{tabular}[c]{@{}c@{}}LSSTA \cite{yang2023long}\end{tabular} &
  \multicolumn{1}{c|}{0.30/0.52} &
  \multicolumn{1}{c|}{0.12/0.20} &
  \multicolumn{1}{c|}{0.28/0.55} &
  \multicolumn{1}{c|}{0.20/0.40} &
  \multicolumn{1}{c|}{0.16/0.32} &
  0.21/0.40 \\ \hline
  \begin{tabular}[c]{@{}c@{}}RCPN \cite{zhu2023reciprocal}\end{tabular} &
  \multicolumn{1}{c|}{0.48/0.86} &
  \multicolumn{1}{c|}{0.38/0.68} &
  \multicolumn{1}{c|}{0.31/0.58} &
  \multicolumn{1}{c|}{0.25/0.44} &
  \multicolumn{1}{c|}{0.23/0.35} &
  0.33/0.58 \\ \hline
  \begin{tabular}[c]{@{}c@{}}STS LSTM \cite{zhang2023spatial}\end{tabular} &
  \multicolumn{1}{c|}{0.46/0.81} &
  \multicolumn{1}{c|}{ 0.20/0.29} &
  \multicolumn{1}{c|}{0.38/0.70} &
  \multicolumn{1}{c|}{0.30/0.57} &
  \multicolumn{1}{c|}{0.24/0.48} &
  0.32/0.57 \\ \hline
\begin{tabular}[c]{@{}c@{}}SIM  \cite{li2023synchronous}\end{tabular} &
  \multicolumn{1}{c|}{0.32/0.53} &
  \multicolumn{1}{c|}{0.32/0.53} &
  \multicolumn{1}{c|}{0.16/0.34} &
  \multicolumn{1}{c|}{0.12/0.25} &
  \multicolumn{1}{c|}{0.09/0.18} &
  0.16/0.29 \\ \hline  
\begin{tabular}[c]{@{}c@{}}SRGAT  \cite{chen2023goal}\end{tabular} &
  \multicolumn{1}{c|}{0.25/0.38} &
  \multicolumn{1}{c|}{0.10/0.15} &
  \multicolumn{1}{c|}{0.21/0.38} &
  \multicolumn{1}{c|}{0.16/0.28} &
  \multicolumn{1}{c|}{0.12/0.21} &
  0.17/0.28 \\ \hline 
\begin{tabular}[c]{@{}c@{}}VNAGT  \cite{chen2023vnagt}\end{tabular} &
  \multicolumn{1}{c|}{0.52/0.88} &
  \multicolumn{1}{c|}{0.16/0.25} &
  \multicolumn{1}{c|}{0.27/0.51} &
  \multicolumn{1}{c|}{0.23/0.44} &
  \multicolumn{1}{c|}{0.18/0.33} &
  0.27/0.48 \\ \hline   
\begin{tabular}[c]{@{}c@{}}MetaTraj w/MemoNet \cite{shi2023metatraj}\end{tabular} &
  \multicolumn{1}{c|}{0.38/0.59} &
  \multicolumn{1}{c|}{0.11/0.16} &
  \multicolumn{1}{c|}{0.22/0.41} &
  \multicolumn{1}{c|}{0.18/0.30} &
  \multicolumn{1}{c|}{0.13/0.26} &
  0.20/0.34 \\ \hline  
\begin{tabular}[c]{@{}c@{}}SOCIAL SAGAN \cite{yang2023social}\end{tabular} &
  \multicolumn{1}{c|}{0.65/1.19} &
  \multicolumn{1}{c|}{0.36/0.70} &
  \multicolumn{1}{c|}{0.54/1.14} &
  \multicolumn{1}{c|}{0.33/0.66} &
  \multicolumn{1}{c|}{0.29/0.61} &
  0.43/0.86 \\ \hline   
\begin{tabular}[c]{@{}c@{}}SSAGCN (B) \cite{lv2023ssagcn}\end{tabular} &
  \multicolumn{1}{c|}{0.21/0.38} &
  \multicolumn{1}{c|}{0.11/0.19} &
  \multicolumn{1}{c|}{0.14/0.25} &
  \multicolumn{1}{c|}{0.12/0.22} &
  \multicolumn{1}{c|}{0.09/0.15} &
  0.13/0.24 \\ \hline
\begin{tabular}[c]{@{}c@{}}Our (B + TWD)\end{tabular} &
  \multicolumn{1}{c|}{0.21/0.38} &
  \multicolumn{1}{c|}{\textbf{0.07/0.10}} &
  \multicolumn{1}{c|}{\textbf{0.10/0.18}} &
  \multicolumn{1}{c|}{\textbf{0.09/0.17}} &
  \multicolumn{1}{c|}{\textbf{0.07/0.11}} &
  \textbf{0.11/0.19} \\ \hline
\begin{tabular}[c]{@{}c@{}}RD($\%$) ADE/FDE \end{tabular} &
  \multicolumn{1}{c|}{-} &
  \multicolumn{1}{c|}{-} &
  \multicolumn{1}{c|}{-} &
  \multicolumn{1}{c|}{-} &
  \multicolumn{1}{c|}{-} &
  \textbf{16.7/23.3} \\ \hline  
\end{tabular}
\end{table*}

\begin{table}[]
\centering
\caption{Quantitative Results on the TrajNet++ dataset using \textit{TWD} during  training. (B) refers to the baseline model. RD($\%$) refers to the relative percent difference with respect to the baseline.}
\label{table_trajnet}
\resizebox{\columnwidth}{!}{%
\begin{tabular}{|c|cc|}
\hline
\textbf{Model} &
  \textbf{\begin{tabular}[c]{@{}l@{}}Scene-level \\ Min ADE ($\downarrow$) \end{tabular}} &
  \textbf{\begin{tabular}[c]{@{}l@{}}Scene-level \\ Min FDE ($\downarrow$) \end{tabular}} \\ \hline
\multicolumn{1}{|c|}{Social gan \cite{gupta2018social}} & \multicolumn{1}{c|}{0.57}          & 1.24          \\ \hline
\multicolumn{1}{|c|}{Social attention \cite{vemula2018social}} & \multicolumn{1}{c|}{0.56}          & 1.21          \\ \hline
\multicolumn{1}{|c|}{Social-bigat \cite{kosaraju2019social} } & \multicolumn{1}{c|}{0.56}          & 1.22         \\ \hline
\multicolumn{1}{|c|}{trajectron \cite{ivanovic2019trajectron} } & \multicolumn{1}{c|}{0.60}          & 1.28          \\ \hline
\multicolumn{1}{|c|}{AIN \cite{zhu2020robust}} & \multicolumn{1}{c|}{0.620}          & 1.240         \\ \hline
\multicolumn{1}{|c|}{PecNet \cite{mangalam2020not} } & \multicolumn{1}{c|}{0.570}          & 1.180         \\ \hline
\multicolumn{1}{|c|}{AMENet \cite{cheng2021amenet}  } & \multicolumn{1}{c|}{0.620}          & 1.300         \\ \hline
\multicolumn{1}{|c|}{socially-aware \cite{saadatnejad2022socially} } & \multicolumn{1}{c|}{0.60}          & 1.28          \\ \hline
\multicolumn{1}{|c|}{Linear Extrapolation \cite{girgis2022latent} } & \multicolumn{1}{c|}{0.409}          & 0.897          \\ \hline
\multicolumn{1}{|c|}{AntiSocial \cite{girgis2022latent} }   & \multicolumn{1}{c|}{0.316}          & 0.632          \\ \hline
\multicolumn{1}{|c|}{Ego \cite{girgis2022latent} }          & \multicolumn{1}{c|}{0.214}          & 0.431          \\ \hline
\multicolumn{1}{|c|}{AutoBot (B) \cite{girgis2022latent} }              & \multicolumn{1}{c|}{0.128}          & 0.234          \\ \hline
\multicolumn{1}{|c|}{Our (B + TWD)}                  & \multicolumn{1}{c|}{\textbf{0.105}} & \textbf{0.195} \\ \hline
\multicolumn{1}{|c|}{\begin{tabular}[c]{@{}c@{}} RD($\%$)\end{tabular}} &
  \multicolumn{1}{c|}{\textbf{19.7}} &
  \textbf{18.2} \\ \hline
\end{tabular}}
\end{table}

\subsection{Quantitative Results} \label{Quantita_result}

\subsubsection{Performance on the NBA Dataset}
The NBA dataset exhibits complex agent interactions, making it an ideal testbed for evaluating our proposed method. Our approach emphasizes explicit temporal learning, which aligns well with the dynamic nature of NBA scenarios. We predict future positions at ten timestamps (4.0 seconds ahead) based on historical trajectories from the preceding five timestamps (2.0 seconds). The summarized results in Table \ref{nba_dataset} present a comparative analysis with nine methods. Our findings demonstrate that the \textit{TWD} approach outperforms other methods significantly. Specifically, the minimum Average Displacement Error (minADE) and minimum Final Displacement Error (minFDE) at 4.0 seconds are reduced to 0.92 and 1.19, respectively, in comparison to GroupNet \cite{Xu_2022_CVPR} (baseline). This reduction represents a substantial relative gain of 20.5\% and 34.7\%, highlighting the effectiveness of our approach.

\subsubsection{Performance on the ETH-UCY Dataset}
 
In this experiment, we conducted a comparison of our results against fifteen methods, and the corresponding ADE and FDE values are detailed in Table~\ref{table_eth}. Significantly, our method exhibits a noteworthy enhancement in predictive accuracy. The proposed \textit{TWD} approach achieves a 16.7\% relative gain in ADE and a 23.3\% relative gain in FDE when compared to the baseline SSAGCN model.

\subsubsection{Performance on the TRAJNET++ Dataset}
In this experiment, we predict the subsequent 12 timesteps for all agents based on the preceding 9 timesteps. Our evaluation focuses on the AutoBot baseline, which features a transformer-based architecture comprising an encoder-decoder network. Importantly, the use of \textit{TWD} demonstrated performance improvements compared to its ablated counterparts, particularly in scene-level metrics, as detailed in Table~\ref{table_trajnet}. Overall, there is a 19.7\% and 18.2\% relative gain in ADE and FDE values, respectively, compared to the baseline AutoBot model.
 \begin{figure*}
    \centering
    \includegraphics[scale=.15]{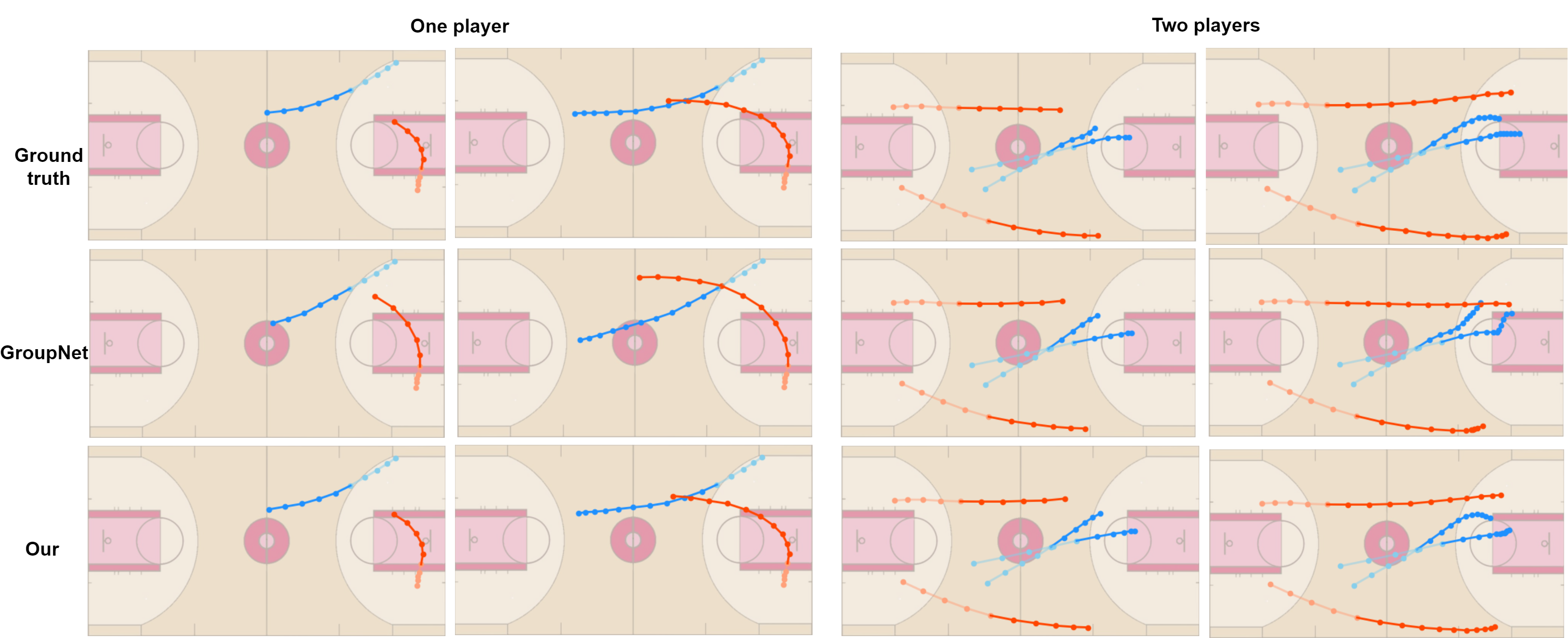}
    \caption{Qualitative results on the NBA dataset. We visualize the trajectories of one player and two players from each team (\textcolor{cyan}{cyan} and \textcolor{red}{red}) in comparison to GroupNet \cite{Xu_2022_CVPR} and ground truth. The light color represents the past trajectory, and the solid color represents the predicted waypoints. In the first and third columns, the model predicts the waypoints for the next five timestamps, while in the second and fourth columns, it forecasts the waypoints for the next ten timestamps.}
    \label{nba_vis}
\end{figure*}

\subsection{Qualitative Results on NBA Dataset}

We extended our investigation through qualitative experiments. Specifically, we conducted visualizations (see Figure~\ref{nba_vis}) of sampled trajectories for one and two players from each team in the NBA dataset. Subsequently, we explored variations in prediction length, forecasting the subsequent five timestamps and forecasting the subsequent ten timestamps, to assess the temporal accuracy of long-term waypoint predictions. The outcomes of these experiments yielded the following insights: i) Our \textit{TWD} method consistently generated more precise trajectories than GroupNet \cite{Xu_2022_CVPR}.
ii) Our method exhibited accurate predictions in a long-time horizon, where GroupNet encountered challenges. This improvement can be attributed to our proposed approach's ability to explicitly learn temporal interactions among players in the scenes. Figure \ref{nba_vis} presents a comparative visualization of predicted trajectories by our method, GroupNet, and ground-truth trajectories on the NBA dataset. 

\subsection{Qualitative Results on ETH-UCY Dataset}

\begin{figure}[b]
    \centering
    \includegraphics[scale=.1]{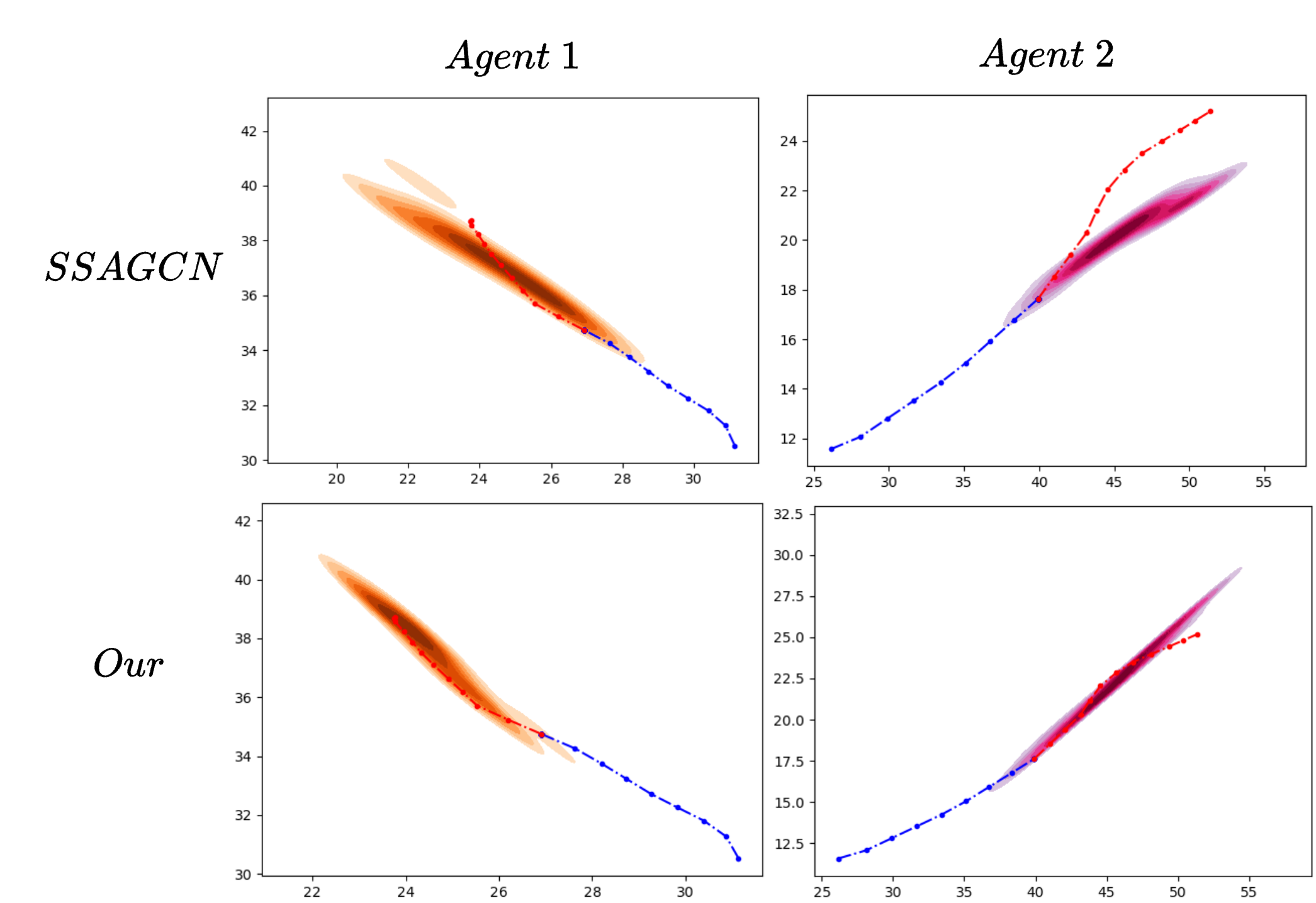}
    \caption{Visualization of temporal density estimations for the agent on the ETH/UCY datasets employing SSAGCN \cite{lv2023ssagcn} and our method. The color-coded density (\textcolor{orange}{orange} for Agent 1 and \textcolor{purple}{purple} for Agent 2) illustrates the predicted distribution of future trajectories. The blue dotted line signifies the past trajectory (8 timestamps), while the red dotted line corresponds to the actual ground truth (12 timestamps).}
    \label{ethvis}
\end{figure}

We have also demonstrated the visualization of predicted density on the ETH-UCY dataset, as shown in Figure \ref{ethvis}. We selected different agents (Agent 1 \& 2) to provide a comprehensive visual representation of our results. The orange (Agent 1) and purple (Agent 2) colors illustrate our method's accurate predictions of future density, effectively capturing the agent's future distribution. Compared to SSAGCN, our approach successfully captures the ground truth trajectories. 

\subsection{Ablation Studies} \label{ablation}

In the following Section, we present ablation studies and analyses to assess the effectiveness of \textit{TWD} under various scenarios. During training, we comprehensively examine the dropping process using stochastic \textit{TWD}. Additionally, we evaluate \textit{TWD} performance when dropping multiple temporal waypoints. Finally, we analyze the significance of our approach by assessing its impact on missing waypoints in the temporal sequence.

\subsubsection{Effect of Using TWD at Training Time} \label{abalation_trainingtesting}
We investigate the impact of incorporating \textit{TWD} during training, and the results are outlined in Table \ref{ttdrop}. Our findings suggest that training the model with stochastic temporal waypoint drops consistently leads to performance enhancements compared to the baseline. 

\subsubsection{Single vs. Multiple Temporal Waypoint Drops}
\label{abalation_multodrop}
Table \ref{abs_mult} shows the ADE/FDE with a single temporal waypoint drop (\textit{D=1}) and multiple temporal waypoint drops (\textit{D=2}) during training. It is worth noting that the best result is achieved when we drop a single waypoint during training. A possible reason may be that, in the case of multiple temporal waypoint drops, the model is not able to effectively learn the underlying temporal representation from the remaining waypoints due to loss of significant information.

\subsubsection{Effect of Missing Waypoints at Test Time} \label{missing}
We also investigate the effects of missing waypoints (i.e., incomplete temporal sequence) on the model performance. We conducted experiments, as presented in Table \ref{table_occulusion}, where we randomly removed the same waypoint from the temporal sequence of each model to simulate missing waypoint scenarios and evaluated the trajectory prediction. These missing waypoints hinder the baseline model's performance and impact its predictions, leading to a significant deterioration in results. In these scenarios, our approach outperforms the baseline significantly. We observe a 65.8\% relative gain in ADE and a 53.5\% relative gain in FDE when compared to the baseline AutoBot model on the TRAJNET++ dataset.

\begin{figure*}[t]
    \centering
    \includegraphics[scale=.15]{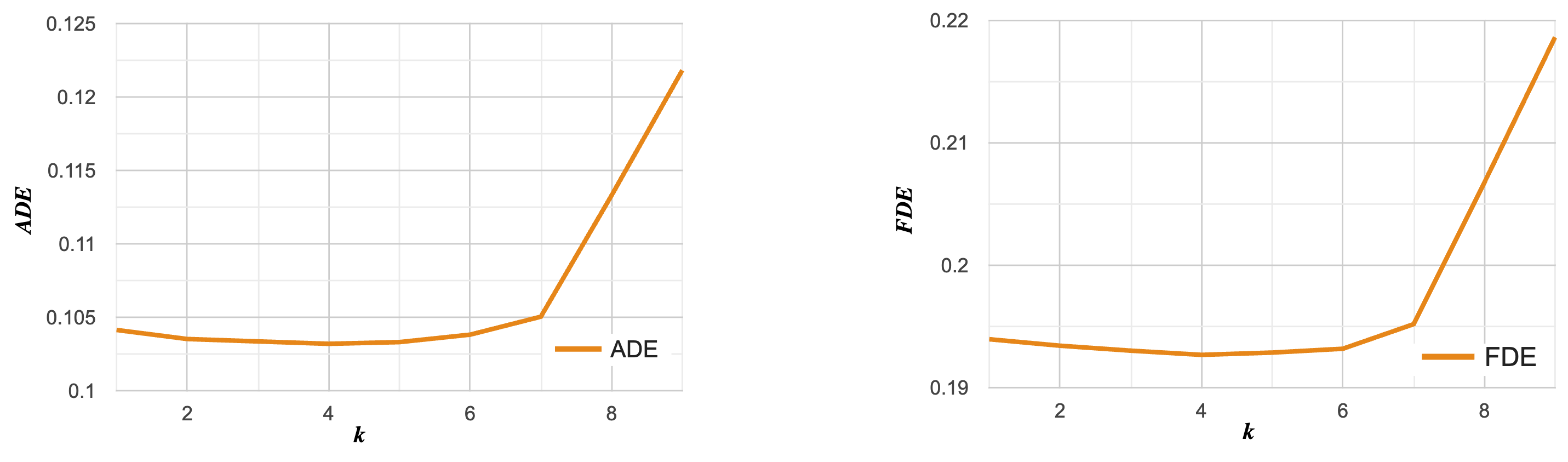}
    \caption{Visualization of ADE/FDE with different fixed waypoint drops at test time, while using stochastic drops during training, shows that an intermediate waypoint dropping using \textit{$k=4$} in the fixed waypoint dropping process at test time yields the best results on the TrajNet++ dataset.}
    \label{plot_traj3}
\end{figure*}

\begin{table}[]
\caption{Comparison of results between using TWD during training versus not using TWD during training.}
\label{ttdrop}
\resizebox{\columnwidth}{!}{%
\begin{tabular}{|c|c|c|c|c|c|}
\hline
Model & Dataset & Training & ADE   & FDE   & ADE/FDE RD($\%$) \\ \hline
\multirow{2}{*}{AutoBot}  & \multirow{2}{*}{TRAJNET++} & $w/o~TWD$ & 0.128 & 0.234 & \multirow{2}{*}{\textbf{19.7/18.2}} \\ \cline{3-5}
               &         & $w/~TWD$    & \textbf{0.105} & \textbf{0.195} &           \\ \hline
\multirow{2}{*}{GroupNet} & \multirow{2}{*}{NBA}       & $w/o~TWD$ & 1.13  & 1.69  & \multirow{2}{*}{\textbf{20.5/34.7}}   \\ \cline{3-5}
               &         & $w/~TWD$    & \textbf{0.92}  & \textbf{1.19}  &           \\ \hline
\multirow{2}{*}{SSAGCN}   & \multirow{2}{*}{ETH-UCY}   & $w/o~TWD$ & 0.13  & 0.24  & \multirow{2}{*}{\textbf{16.7/23.3}} \\ \cline{3-5}
               &         & $w/~TWD$    & \textbf{0.11} & \textbf{0.19} &           \\ \hline
\end{tabular}%
}
\end{table}

\begin{table}[]
\caption{Results for \textit{GroupNet+TWD} model using a single temporal waypoint drop (\textit{D=1}) and multiple temporal waypoint drops (\textit{D=2}) during training on the NBA dataset.}
\label{abs_mult}
\resizebox{\columnwidth}{!}{%
\begin{tabular}{|c|cccccccc|}
\hline
\multirow{2}{*}{\begin{tabular}[c]{@{}c@{}}\textit{D} \\ \end{tabular}} &
  \multicolumn{8}{c|}{Validation Accuracy} \\ \cline{2-9} 
 &
  \multicolumn{4}{c|}{ADE} &
  \multicolumn{4}{c|}{FDE} \\ \hline
Time &
  \multicolumn{1}{c|}{1.0s} &
  \multicolumn{1}{c|}{2.0s} &
  \multicolumn{1}{c|}{3.0s} &
  \multicolumn{1}{c|}{4.0s} &
  \multicolumn{1}{c|}{1.0s} &
  \multicolumn{1}{c|}{2.0s} &
  \multicolumn{1}{c|}{3.0s} &
  4.0s \\ \hline
\textit{D=1} &
  \multicolumn{1}{c|}{\textbf{0.246}} &
  \multicolumn{1}{c|}{\textbf{0.462}} &
  \multicolumn{1}{c|}{\textbf{0.692}} &
  \multicolumn{1}{c|}{\textbf{0.920}} &
  \multicolumn{1}{c|}{\textbf{0.320}} &
  \multicolumn{1}{c|}{\textbf{0.656}} &
  \multicolumn{1}{c|}{\textbf{0.954}} &
  \textbf{1.185} \\ \hline
\textit{D=2} &
  \multicolumn{1}{c|}{0.303} &
  \multicolumn{1}{c|}{0.545} &
  \multicolumn{1}{c|}{0.791} &
  \multicolumn{1}{c|}{1.032} &
  \multicolumn{1}{c|}{0.394} &
  \multicolumn{1}{c|}{0.766} &
  \multicolumn{1}{c|}{1.072} &
  1.320 \\ \hline
\end{tabular}%
}
\end{table}

\begin{table}[]
\caption{Results of experiments involving missing waypoints in the temporal sequence at test time show that \textit{TWD} demonstrates robustness, while baseline models experience substantial performance degradation. (B) refers to the baseline model. RD($\%$) refers to the relative percent difference with respect to the baseline.}
\label{table_occulusion}
\resizebox{\columnwidth}{!}{%
\begin{tabular}{|c|c|c|c|c|}
\hline
Methods        & Environment                        & ADE   & FDE   & ADE/FDE RD($\%$)              \\ \hline
AutoBot (B)       & Missing Waypoints & 0.210 & 0.341 & \multirow{2}{*}{\textbf{65.8/53.5}} \\ \cline{1-1} \cline{3-4}
AutoBot + TWD &   TRAJNET++                                  & \textbf{0.106} & \textbf{0.197} &                            \\ \hline
GroupNet (B)      & Missing Waypoints & 1.70  & 2.36  & \multirow{2}{*}{\textbf{55.6/62.2}} \\ \cline{1-1} \cline{3-4}
GroupNet + TWD &    NBA                                 & \textbf{0.96}  & \textbf{1.24}  &                            \\ \hline
SSAGCN (B)      & Missing Waypoints & 0.173 & 0.269 & \multirow{2}{*}{\textbf{41.1/33.4}} \\ \cline{1-1} \cline{3-4}
SSAGCN + TWD   &     ETH-UCY                               & \textbf{0.114} & \textbf{0.192} &                            \\ \hline
\end{tabular}%
}
\end{table}

\begin{table}[]
\centering
\caption{Results of $S_d$/$F_d$ drops at test time while using stochastic drops during training. Where $S_d$ and $F_d$ denote stochastic and fixed drop, respectively. The value of $k$ represents the waypoint being dropped in fixed drop during testing. (B) denotes the baseline model.}
\label{drop}
\resizebox{\columnwidth}{!}{%
\begin{tabular}{|c|c|c|c|c|c|c|}
\hline
Model        & Dataset   & Train & Test  & \textit{k} & ADE            & FDE            \\ \hline
AutoBot (B)               & TRAJNET++ & -     & -     & -            & 0.128          & 0.234          \\ \hline
AutoBot+TWD               & TRAJNET++ & $S_d$ & -     & -            & 0.105          & 0.195         \\ \hline
AutoBot+TWD   & TRAJNET++ & $S_d$ & $S_d$ & -            & 0.106          & 0.197          \\ \hline
AutoBot+TWD   & TRAJNET++ & $S_d$ & $F_d$ & 4            & \textbf{0.103} & \textbf{0.192} \\ \hline

\end{tabular}
}
\end{table}

\subsubsection{Temporal Waypoint Dropping at Test Time} \label{fixed}

We also explore temporal waypoint dropping at test time using stochastic dropping \textit{($S_d$)} and fixed dropping \textit{($F_d$)}. The fixed process identifies the waypoint that yields the maximum evaluation score on the validation set (Equation~\ref{eqsix}) while being dropped from the temporal sequence.
Assuming $k^{th}$ waypoint drop maximizes the evaluation scores, then we will calculate $TWD(\omega)$ using that $k$ value. $TWD(\omega)$ is passed as input to the model \(g_{\theta}(.)\) to generate future predictions.

The strategy optimizes the following objective function:

\begin{equation} \label{eqsix}
\begin{aligned}
    k = \arg\max_{1 \le k \le |X_i|} \sum_{i=1}^{N_{}^{}}D(g_{\theta}(S_i^k),Y_i)
\end{aligned}
\end{equation}

where \(g_{\theta}(.)\) represents the prediction model that generates predicted future trajectory for $S_{i}^{k}$, and \(D\) is the quality evaluation metric (e.g., Average Distance Error - ADE, Final Displacement Error - FDE, etc.) used to assess the quality of the generated trajectory. Thus, the above Equation identifies the $k$ value (for dropping the waypoint), which maximizes the evaluation scores on the validation set.

In these experiments, we use stochastic dropping \textit{($S_d$)} in the training phase and use stochastic dropping \textit{($S_d$)} or fixed dropping \textit{($F_d$)} in the testing phase. During training, the model $g_{\theta}(.)$ uses trajectory sequences generated by $TWD(\omega)$ as input. $TWD(\omega)$ is generated by dropping temporal waypoints using a stochastic process, which is then used in the training stage. Please note that in each training iteration, \textit{k} value is randomly chosen and different in $TWD(\omega)$. During testing, the model $g_{\theta}(.)$ uses trajectory sequences generated by $TWD(\omega)$ either using fixed dropping process \textit{($F_d$)} or using stochastic dropping \textit{($S_d$)} as input for predicting future trajectories $\tau$. 
In a fixed drop process, $TWD(\omega)$ is generated by dropping temporal waypoints using a \textit{k} value chosen using Equation~\ref{eqsix}.

Figure~\ref{plot_traj3} shows the ADE/FDE with different fixed waypoint drops at test time while using stochastic drops during training. It's worth noting that, for the TrajNet++ dataset, the optimal result is achieved with fixed \textit{TWD} with \textit{$k=4$} during testing as shown in Table~\ref{drop}. It is also clear from Table~\ref{drop} that if we use stochastic drops during both training and testing, then its accuracy is lower than the variant where we only use stochastic drops during training. Therefore, we can conclude that using stochastic drops during training improves the accuracy significantly as compared to the baseline, and using fixed drops during testing leads to a minor improvement in accuracy. Nevertheless, predicting the \textit{$k$} value in fixed drops for real-time applications is not always feasible, as it may change depending on the dynamic environment and agent behavior.

\section{Conclusion}
This paper introduces a novel approach (TWD) to explicitly capture temporal dependencies within trajectory prediction tasks. TWD leverages stochastic processes to select and drop waypoints from temporal sequences. We conducted experiments using three trajectory prediction benchmark datasets with various baseline architectures to validate our novel approach. Our extensive experiments demonstrate the substantial performance gains achieved by integrating TWD into these baseline architectures. Temporal waypoint dropping facilitates the learning of underlying temporal patterns and significantly enhances predictions in dynamic environments where the observer's temporal sequence may be missing.

\bibliographystyle{IEEEtran}
\bibliography{ref}

\begin{thebibliography}{10}
\providecommand{\url}[1]{#1}
\csname url@samestyle\endcsname
\providecommand{\newblock}{\relax}
\providecommand{\bibinfo}[2]{#2}
\providecommand{\BIBentrySTDinterwordspacing}{\spaceskip=0pt\relax}
\providecommand{\BIBentryALTinterwordstretchfactor}{4}
\providecommand{\BIBentryALTinterwordspacing}{\spaceskip=\fontdimen2\font plus
\BIBentryALTinterwordstretchfactor\fontdimen3\font minus \fontdimen4\font\relax}
\providecommand{\BIBforeignlanguage}[2]{{%
\expandafter\ifx\csname l@#1\endcsname\relax
\typeout{** WARNING: IEEEtran.bst: No hyphenation pattern has been}%
\typeout{** loaded for the language `#1'. Using the pattern for}%
\typeout{** the default language instead.}%
\else
\language=\csname l@#1\endcsname
\fi
#2}}
\providecommand{\BIBdecl}{\relax}
\BIBdecl

\bibitem{shi2021sgcn}
L.~Shi, L.~Wang, C.~Long, S.~Zhou, M.~Zhou, Z.~Niu, and G.~Hua, ``Sgcn: Sparse graph convolution network for pedestrian trajectory prediction,'' in \emph{Proceedings of the IEEE/CVF Conference on Computer Vision and Pattern Recognition}, 2021, pp. 8994--9003.

\bibitem{mohamed2020social}
A.~Mohamed, K.~Qian, M.~Elhoseiny, and C.~Claudel, ``Social-stgcnn: A social spatio-temporal graph convolutional neural network for human trajectory prediction,'' in \emph{Proceedings of the IEEE/CVF conference on computer vision and pattern recognition}, 2020, pp. 14\,424--14\,432.

\bibitem{sekhon2021scan}
J.~Sekhon and C.~Fleming, ``Scan: A spatial context attentive network for joint multi-agent intent prediction,'' in \emph{Proceedings of the AAAI Conference on Artificial Intelligence}, vol.~35, no.~7, 2021, pp. 6119--6127.

\bibitem{gupta2018social}
A.~Gupta, J.~Johnson, L.~Fei-Fei, S.~Savarese, and A.~Alahi, ``Social gan: Socially acceptable trajectories with generative adversarial networks,'' in \emph{Proceedings of the IEEE conference on computer vision and pattern recognition}, 2018, pp. 2255--2264.

\bibitem{cao2018brits}
W.~Cao, D.~Wang, J.~Li, H.~Zhou, L.~Li, and Y.~Li, ``Brits: Bidirectional recurrent imputation for time series,'' \emph{Advances in neural information processing systems}, vol.~31, 2018.

\bibitem{bai2018empirical}
S.~Bai, J.~Z. Kolter, and V.~Koltun, ``An empirical evaluation of generic convolutional and recurrent networks for sequence modeling,'' \emph{arXiv preprint arXiv:1803.01271}, 2018.

\bibitem{vaswani2017attention}
A.~Vaswani, N.~Shazeer, N.~Parmar, J.~Uszkoreit, L.~Jones, A.~N. Gomez, {\L}.~Kaiser, and I.~Polosukhin, ``Attention is all you need,'' \emph{Advances in neural information processing systems}, vol.~30, 2017.

\bibitem{Xu_2022_CVPR}
C.~Xu, M.~Li, Z.~Ni, Y.~Zhang, and S.~Chen, ``Groupnet: Multiscale hypergraph neural networks for trajectory prediction with relational reasoning,'' in \emph{Proceedings of the IEEE/CVF Conference on Computer Vision and Pattern Recognition (CVPR)}, June 2022, pp. 6498--6507.

\bibitem{lv2023ssagcn}
P.~Lv, W.~Wang, Y.~Wang, Y.~Zhang, M.~Xu, and C.~Xu, ``Ssagcn: social soft attention graph convolution network for pedestrian trajectory prediction,'' \emph{IEEE transactions on neural networks and learning systems}, 2023.

\bibitem{girgis2022latent}
\BIBentryALTinterwordspacing
R.~Girgis, F.~Golemo, F.~Codevilla, M.~Weiss, J.~A. D'Souza, S.~E. Kahou, F.~Heide, and C.~Pal, ``Latent variable sequential set transformers for joint multi-agent motion prediction,'' in \emph{International Conference on Learning Representations}, 2022. [Online]. Available: \url{https://openreview.net/forum?id=Dup_dDqkZC5}
\BIBentrySTDinterwordspacing

\bibitem{pellegrini2009you}
S.~Pellegrini, A.~Ess, K.~Schindler, and L.~Van~Gool, ``You'll never walk alone: Modeling social behavior for multi-target tracking,'' in \emph{2009 IEEE 12th international conference on computer vision}.\hskip 1em plus 0.5em minus 0.4em\relax IEEE, 2009, pp. 261--268.

\bibitem{lerner2007crowds}
A.~Lerner, Y.~Chrysanthou, and D.~Lischinski, ``Crowds by example,'' in \emph{Computer graphics forum}, vol.~26, no.~3.\hskip 1em plus 0.5em minus 0.4em\relax Wiley Online Library, 2007, pp. 655--664.

\bibitem{zhan2018generating}
E.~Zhan, S.~Zheng, Y.~Yue, L.~Sha, and P.~Lucey, ``Generating multi-agent trajectories using programmatic weak supervision,'' \emph{arXiv preprint arXiv:1803.07612}, 2018.

\bibitem{Kothari2020HumanTF}
P.~Kothari, S.~Kreiss, and A.~Alahi, ``Human trajectory forecasting in crowds: A deep learning perspective,'' \emph{IEEE Transactions on Intelligent Transportation Systems}, pp. 1--15, 2021.

\bibitem{helbing1995social}
D.~Helbing and P.~Molnar, ``Social force model for pedestrian dynamics,'' \emph{Physical review E}, vol.~51, no.~5, p. 4282, 1995.

\bibitem{salzmann2020trajectron++}
T.~Salzmann, B.~Ivanovic, P.~Chakravarty, and M.~Pavone, ``Trajectron++: Dynamically-feasible trajectory forecasting with heterogeneous data,'' in \emph{Computer Vision--ECCV 2020: 16th European Conference, Glasgow, UK, August 23--28, 2020, Proceedings, Part XVIII 16}.\hskip 1em plus 0.5em minus 0.4em\relax Springer, 2020, pp. 683--700.

\bibitem{park2020diverse}
S.~H. Park, G.~Lee, J.~Seo, M.~Bhat, M.~Kang, J.~Francis, A.~Jadhav, P.~P. Liang, and L.-P. Morency, ``Diverse and admissible trajectory forecasting through multimodal context understanding,'' in \emph{Computer Vision--ECCV 2020: 16th European Conference, Glasgow, UK, August 23--28, 2020, Proceedings, Part XI 16}.\hskip 1em plus 0.5em minus 0.4em\relax Springer, 2020, pp. 282--298.

\bibitem{lee2022muse}
M.~Lee, S.~S. Sohn, S.~Moon, S.~Yoon, M.~Kapadia, and V.~Pavlovic, ``Muse-vae: multi-scale vae for environment-aware long term trajectory prediction,'' in \emph{Proceedings of the IEEE/CVF Conference on Computer Vision and Pattern Recognition}, 2022, pp. 2221--2230.

\bibitem{xu2022dynamic}
C.~Xu, Y.~Wei, B.~Tang, S.~Yin, Y.~Zhang, and S.~Chen, ``Dynamic-group-aware networks for multi-agent trajectory prediction with relational reasoning,'' \emph{arXiv preprint arXiv:2206.13114}, 2022.

\bibitem{mangalam2020not}
K.~Mangalam, H.~Girase, S.~Agarwal, K.-H. Lee, E.~Adeli, J.~Malik, and A.~Gaidon, ``It is not the journey but the destination: Endpoint conditioned trajectory prediction,'' in \emph{Computer Vision--ECCV 2020: 16th European Conference, Glasgow, UK, August 23--28, 2020, Proceedings, Part II 16}.\hskip 1em plus 0.5em minus 0.4em\relax Springer, 2020, pp. 759--776.

\bibitem{sophie19}
A.~Sadeghian, V.~Kosaraju, A.~Sadeghian, N.~Hirose, H.~Rezatofighi, and S.~Savarese, ``Sophie: An attentive gan for predicting paths compliant to social and physical constraints,'' in \emph{IEEE Conference on Computer Vision and Pattern Recognition (CVPR)}, no. CONF, 2019.

\bibitem{hu2020collaborative}
Y.~Hu, S.~Chen, Y.~Zhang, and X.~Gu, ``Collaborative motion prediction via neural motion message passing,'' in \emph{Proceedings of the IEEE/CVF conference on computer vision and pattern recognition}, 2020, pp. 6319--6328.

\bibitem{kosaraju2019social}
V.~Kosaraju, A.~Sadeghian, R.~Mart{\'\i}n-Mart{\'\i}n, I.~Reid, H.~Rezatofighi, and S.~Savarese, ``Social-bigat: Multimodal trajectory forecasting using bicycle-gan and graph attention networks,'' \emph{Advances in Neural Information Processing Systems}, vol.~32, 2019.

\bibitem{mao2023leapfrog}
W.~Mao, C.~Xu, Q.~Zhu, S.~Chen, and Y.~Wang, ``Leapfrog diffusion model for stochastic trajectory prediction,'' in \emph{Proceedings of the IEEE/CVF Conference on Computer Vision and Pattern Recognition}, 2023, pp. 5517--5526.

\bibitem{shi2023representing}
L.~Shi, L.~Wang, C.~Long, S.~Zhou, W.~Tang, N.~Zheng, and G.~Hua, ``Representing multimodal behaviors with mean location for pedestrian trajectory prediction,'' \emph{IEEE transactions on pattern analysis and machine intelligence}, 2023.

\bibitem{gu2022stochastic}
T.~Gu, G.~Chen, J.~Li, C.~Lin, Y.~Rao, J.~Zhou, and J.~Lu, ``Stochastic trajectory prediction via motion indeterminacy diffusion,'' in \emph{Proceedings of the IEEE/CVF Conference on Computer Vision and Pattern Recognition}, 2022, pp. 17\,113--17\,122.

\bibitem{yuan2021agentformer}
Y.~Yuan, X.~Weng, Y.~Ou, and K.~M. Kitani, ``Agentformer: Agent-aware transformers for socio-temporal multi-agent forecasting,'' in \emph{Proceedings of the IEEE/CVF International Conference on Computer Vision}, 2021, pp. 9813--9823.

\bibitem{zhou2023query}
Z.~Zhou, J.~Wang, Y.-H. Li, and Y.-K. Huang, ``Query-centric trajectory prediction,'' in \emph{Proceedings of the IEEE/CVF Conference on Computer Vision and Pattern Recognition}, 2023, pp. 17\,863--17\,873.

\bibitem{yu2020spatio}
C.~Yu, X.~Ma, J.~Ren, H.~Zhao, and S.~Yi, ``Spatio-temporal graph transformer networks for pedestrian trajectory prediction,'' in \emph{Computer Vision--ECCV 2020: 16th European Conference, Glasgow, UK, August 23--28, 2020, Proceedings, Part XII 16}.\hskip 1em plus 0.5em minus 0.4em\relax Springer, 2020, pp. 507--523.

\bibitem{chiara2022goal}
L.~F. Chiara, P.~Coscia, S.~Das, S.~Calderara, R.~Cucchiara, and L.~Ballan, ``Goal-driven self-attentive recurrent networks for trajectory prediction,'' in \emph{Proceedings of the IEEE/CVF Conference on Computer Vision and Pattern Recognition}, 2022, pp. 2518--2527.

\bibitem{bae2023set}
I.~Bae and H.-G. Jeon, ``A set of control points conditioned pedestrian trajectory prediction,'' in \emph{Proceedings of the AAAI Conference on Artificial Intelligence}, vol.~37, no.~5, 2023, pp. 6155--6165.

\bibitem{chen2023goal}
X.~Chen, F.~Luo, F.~Zhao, and Q.~Ye, ``Goal-guided and interaction-aware state refinement graph attention network for multi-agent trajectory prediction,'' \emph{IEEE Robotics and Automation Letters}, vol.~9, no.~1, pp. 57--64, 2023.

\bibitem{pmlr-v80-kipf18a}
\BIBentryALTinterwordspacing
T.~Kipf, E.~Fetaya, K.-C. Wang, M.~Welling, and R.~Zemel, ``Neural relational inference for interacting systems,'' in \emph{Proceedings of the 35th International Conference on Machine Learning}, ser. Proceedings of Machine Learning Research, J.~Dy and A.~Krause, Eds., vol.~80.\hskip 1em plus 0.5em minus 0.4em\relax PMLR, 10--15 Jul 2018, pp. 2688--2697. [Online]. Available: \url{https://proceedings.mlr.press/v80/kipf18a.html}
\BIBentrySTDinterwordspacing

\bibitem{xu2023dynamic}
C.~Xu, Y.~Wei, B.~Tang, S.~Yin, Y.~Zhang, S.~Chen, and Y.~Wang, ``Dynamic-group-aware networks for multi-agent trajectory prediction with relational reasoning,'' \emph{Neural Networks}, 2023.

\bibitem{wang2023trajectory}
R.~Wang, Z.~Hu, X.~Song, and W.~Li, ``Trajectory distribution aware graph convolutional network for trajectory prediction considering spatio-temporal interactions and scene information,'' \emph{IEEE Transactions on Knowledge and Data Engineering}, 2023.

\bibitem{mendieta2021carpe}
M.~Mendieta and H.~Tabkhi, ``Carpe posterum: A convolutional approach for real-time pedestrian path prediction,'' in \emph{Proceedings of the AAAI Conference on Artificial Intelligence}, vol.~35, no.~3, 2021, pp. 2346--2354.

\bibitem{cao2023discovering}
D.~I. S.-T. L.~R. to~Explain Human~Actions, ``Discovering intrinsic spatial-temporal logic rules to explain human actions,'' \emph{Advances in Neural Information Processing Systems}, 2023.

\bibitem{sun2023modality}
J.~Sun, Y.~Li, L.~Chai, and C.~Lu, ``Modality exploration, retrieval and adaptation for trajectory prediction,'' \emph{IEEE Transactions on Pattern Analysis and Machine Intelligence}, 2023.

\bibitem{wong2023msn}
C.~Wong, B.~Xia, Q.~Peng, W.~Yuan, and X.~You, ``Msn: multi-style network for trajectory prediction,'' \emph{IEEE Transactions on Intelligent Transportation Systems}, 2023.

\bibitem{shi2023metatraj}
X.~Shi, H.~Zhang, W.~Yuan, and R.~Shibasaki, ``Metatraj: meta-learning for cross-scene cross-object trajectory prediction,'' \emph{IEEE Transactions on Intelligent Transportation Systems}, 2023.

\bibitem{ye2022bootstrap}
M.~Ye, J.~Xu, X.~Xu, T.~Wang, T.~Cao, and Q.~Chen, ``Bootstrap motion forecasting with self-consistent constraints,'' in \emph{Proceedings of the IEEE/CVF International Conference on Computer Vision}, 2023, pp. 8504--8514.

\bibitem{zhu2020robust}
Y.~Zhu, D.~Ren, M.~Fan, D.~Qian, X.~Li, and H.~Xia, ``Robust trajectory forecasting for multiple intelligent agents in dynamic scene,'' \emph{arXiv preprint arXiv:2005.13133}, 2020.

\bibitem{aydemir2023adapt}
G.~Aydemir, A.~K. Akan, and F.~G{\"u}ney, ``Adapt: Efficient multi-agent trajectory prediction with adaptation,'' \emph{arXiv preprint arXiv:2307.14187}, 2023.

\bibitem{alahi2016social}
A.~Alahi, K.~Goel, V.~Ramanathan, A.~Robicquet, L.~Fei-Fei, and S.~Savarese, ``Social lstm: Human trajectory prediction in crowded spaces,'' in \emph{Proceedings of the IEEE conference on computer vision and pattern recognition}, 2016, pp. 961--971.

\bibitem{huang2019stgat}
Y.~Huang, H.~Bi, Z.~Li, T.~Mao, and Z.~Wang, ``Stgat: Modeling spatial-temporal interactions for human trajectory prediction,'' in \emph{Proceedings of the IEEE/CVF international conference on computer vision}, 2019, pp. 6272--6281.

\bibitem{zhang2023spatial}
C.~Zhang, Z.~Ni, and C.~Berger, ``Spatial-temporal-spectral lstm: A transferable model for pedestrian trajectory prediction,'' \emph{IEEE Transactions on Intelligent Vehicles}, 2023.

\bibitem{vemula2018social}
A.~Vemula, K.~Muelling, and J.~Oh, ``Social attention: Modeling attention in human crowds,'' in \emph{2018 IEEE international Conference on Robotics and Automation (ICRA)}.\hskip 1em plus 0.5em minus 0.4em\relax IEEE, 2018, pp. 4601--4607.

\bibitem{ivanovic2019trajectron}
B.~Ivanovic and M.~Pavone, ``The trajectron: Probabilistic multi-agent trajectory modeling with dynamic spatiotemporal graphs,'' in \emph{Proceedings of the IEEE/CVF International Conference on Computer Vision}, 2019, pp. 2375--2384.

\bibitem{zhong2023visual}
X.~Zhong, X.~Yan, Z.~Yang, W.~Huang, K.~Jiang, R.~W. Liu, and Z.~Wang, ``Visual exposes you: Pedestrian trajectory prediction meets visual intention,'' \emph{IEEE Transactions on Intelligent Transportation Systems}, 2023.

\bibitem{chen2023vnagt}
X.~Chen, H.~Zhang, Y.~Hu, J.~Liang, and H.~Wang, ``Vnagt: Variational non-autoregressive graph transformer network for multi-agent trajectory prediction,'' \emph{IEEE Transactions on Vehicular Technology}, 2023.

\bibitem{yang2023long}
C.~Yang and Z.~Pei, ``Long-short term spatio-temporal aggregation for trajectory prediction,'' \emph{IEEE Transactions on Intelligent Transportation Systems}, vol.~24, no.~4, pp. 4114--4126, 2023.

\bibitem{li2023synchronous}
Y.~Li, C.~Xie, R.~Liang, J.~Du, J.~Zhou, and X.~Li, ``A synchronous bi-directional framework with temporally dependent interaction modeling for pedestrian trajectory prediction,'' \emph{IEEE Transactions on Network Science and Engineering}, 2023.

\bibitem{sun2023stimulus}
J.~Sun, Y.~Li, L.~Chai, and C.~Lu, ``Stimulus verification is a universal and effective sampler in multi-modal human trajectory prediction,'' in \emph{Proceedings of the IEEE/CVF Conference on Computer Vision and Pattern Recognition}, 2023, pp. 22\,014--22\,023.

\bibitem{yang2021novel}
B.~Yang, G.~Yan, P.~Wang, C.-Y. Chan, X.~Song, and Y.~Chen, ``A novel graph-based trajectory predictor with pseudo-oracle,'' \emph{IEEE transactions on neural networks and learning systems}, vol.~33, no.~12, pp. 7064--7078, 2021.

\bibitem{shafiee2021introvert}
N.~Shafiee, T.~Padir, and E.~Elhamifar, ``Introvert: Human trajectory prediction via conditional 3d attention,'' in \emph{Proceedings of the IEEE/cvf Conference on Computer Vision and Pattern recognition}, 2021, pp. 16\,815--16\,825.

\bibitem{pang2021trajectory}
B.~Pang, T.~Zhao, X.~Xie, and Y.~N. Wu, ``Trajectory prediction with latent belief energy-based model,'' in \emph{Proceedings of the IEEE/CVF Conference on Computer Vision and Pattern Recognition}, 2021, pp. 11\,814--11\,824.

\bibitem{mangalam2021goals}
K.~Mangalam, Y.~An, H.~Girase, and J.~Malik, ``From goals, waypoints \& paths to long term human trajectory forecasting,'' in \emph{Proceedings of the IEEE/CVF International Conference on Computer Vision}, 2021, pp. 15\,233--15\,242.

\bibitem{zhu2023reciprocal}
W.~Zhu, Y.~Liu, M.~Zhang, and Y.~Yi, ``Reciprocal consistency prediction network for multi-step human trajectory prediction,'' \emph{IEEE Transactions on Intelligent Transportation Systems}, 2023.

\bibitem{yang2023social}
C.~Yang, H.~Pan, W.~Sun, and H.~Gao, ``Social self-attention generative adversarial networks for human trajectory prediction,'' \emph{IEEE Transactions on Artificial Intelligence}, 2023.

\bibitem{cheng2021amenet}
H.~Cheng, W.~Liao, M.~Y. Yang, B.~Rosenhahn, and M.~Sester, ``Amenet: Attentive maps encoder network for trajectory prediction,'' \emph{ISPRS Journal of Photogrammetry and Remote Sensing}, vol. 172, pp. 253--266, 2021.

\bibitem{saadatnejad2022socially}
S.~Saadatnejad, M.~Bahari, P.~Khorsandi, M.~Saneian, S.-M. Moosavi-Dezfooli, and A.~Alahi, ``Are socially-aware trajectory prediction models really socially-aware?'' \emph{Transportation research part C: emerging technologies}, vol. 141, p. 103705, 2022.

\end{thebibliography}

\vfill

\end{document}